\documentclass[nohyperref]{article}

\usepackage{microtype}
\usepackage{graphicx}
\usepackage{subcaption}
\usepackage{booktabs} 

\usepackage{hyperref}

\usepackage[accepted]{icml2022}

\usepackage{hyperref}
\usepackage{url}
\usepackage{amsmath,amssymb,amsfonts}
\usepackage{amsthm}
\usepackage{thmtools}
\usepackage{float}
\usepackage{pifont}
\newcommand{\indep}{\perp \!\!\! \perp}
\usepackage{amsthm}
\usepackage{algorithm}
\usepackage{algorithmic}
\usepackage{makecell}
\usepackage{wrapfig}
\usepackage{enumitem}
\usepackage{marvosym}

\usepackage{color-edits}
\addauthor{sw}{blue}

\usepackage{xcolor}
\definecolor{expert}{HTML}{008000}
\definecolor{error}{HTML}{f96565}

\newcommand{\norm}[1]{\left\lVert #1 \right\rVert}
\usepackage{tikz}
\usetikzlibrary{arrows,calc,bayesnet} 
\newcommand{\tikzAngleOfLine}{\tikz@AngleOfLine}
\def\tikz@AngleOfLine(#1)(#2)#3{%
\pgfmathanglebetweenpoints{%
\pgfpointanchor{#1}{center}}{%
\pgfpointanchor{#2}{center}}
\pgfmathsetmacro{#3}{\pgfmathresult}%
}

\newcommand{\eref}[1]{(\ref{#1})}
\newcommand{\sref}[1]{Sec. \ref{#1}}
\newcommand{\figref}[1]{Fig. \ref{#1}}

\usepackage[capitalize,noabbrev]{cleveref}

\theoremstyle{plain}
\newtheorem{theorem}{Theorem}[section]

\newtheorem{lemma}[theorem]{Lemma}

\theoremstyle{definition}

\theoremstyle{remark}

\declaretheoremstyle[
headfont=\normalfont\itshape,
qed=\qedsymbol,
]{mypf}

\usepackage[textsize=tiny]{todonotes}

\icmltitlerunning{Causal Imitation Learning under Temporally Correlated Noise}

\begin{document}

\twocolumn[
\icmltitle{Causal Imitation Learning under Temporally Correlated Noise}




\begin{icmlauthorlist}
\icmlauthor{Gokul Swamy}{cmu}
\icmlauthor{Sanjiban Choudhury}{cor}
\icmlauthor{J. Andrew Bagnell}{aur,cmu}
\icmlauthor{Zhiwei Steven Wu}{cmu}
\end{icmlauthorlist}

\icmlaffiliation{cmu}{Carnegie Mellon University}
\icmlaffiliation{aur}{Aurora Innovation}
\icmlaffiliation{cor}{Cornell University}

\icmlcorrespondingauthor{Gokul Swamy}{gswamy@cmu.edu}

\icmlkeywords{Machine Learning, ICML}

\vskip 0.3in
]



\printAffiliationsAndNotice{} 

\begin{abstract}
We develop algorithms for imitation learning from policy data that was corrupted by temporally correlated noise in expert actions. When noise affects multiple timesteps of recorded data, it can manifest as spurious correlations between states and actions that a learner might latch on to, leading to poor policy performance. To break up these spurious correlations, we apply modern variants of the \textit{instrumental variable regression} (IVR) technique of econometrics, enabling us to recover the underlying policy \textit{without} requiring access to an interactive expert. In particular, we present two techniques, one of a generative-modeling flavor (\texttt{DoubIL}) that can utilize access to a simulator, and one of a game-theoretic flavor (\texttt{ResiduIL}) that can be run entirely offline. We find both of our algorithms compare favorably to behavioral cloning on simulated control tasks.
\end{abstract}

\begin{figure*}[t]
\centering
  \includegraphics[width=1.95\columnwidth]{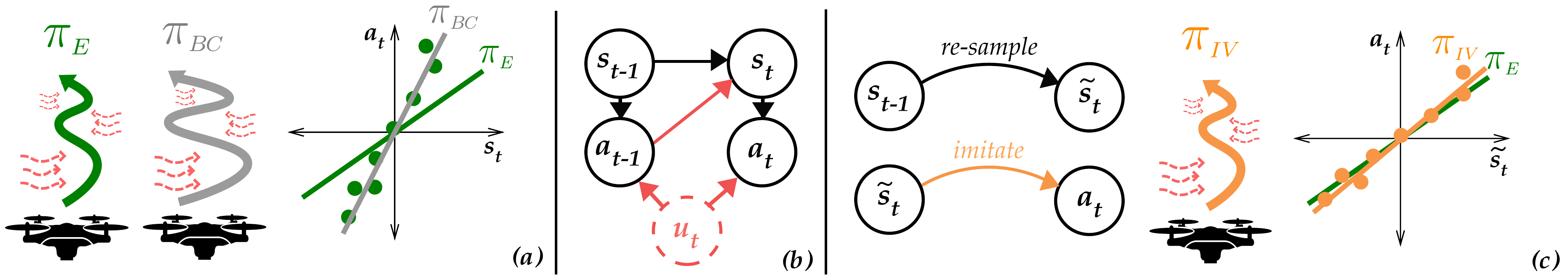}
  \caption{
  \textit{(a)} When temporally correlated noise (e.g. wind) affects expert actions, standard imitation learning approaches like behavioral cloning can amplify this noise, leading to poor test-time performance. \textit{(b)} TCN $u_t$ affects both the input ($s_t$) and output ($a_t$) of our learning procedure. This breaks a cardinal assumption of regression-based approaches like behavioral cloning, rendering them inconsistent. \textit{(c)} We can re-simulate state transitions from a past state, producing fresh samples ($\widetilde{s_t}$). We can then regress from these sampled states to observed expert actions to recover the expert's policy as the noise on inputs and outputs is no longer correlated.}
\label{fig:ffig}
\end{figure*}

\section{Introduction}
\label{sec:introduction}
Much of the theory of imitation learning (IL) tells us that with enough demonstrations, we should be able to accurately recover the expert's policy. A long line of work \cite{ross2011reduction, sun2019provably, spencer2021feedback, swamy2021moments} has derived performance bounds that seem to imply that if infinite-sample training error is driven to zero, value equivalence to the expert policy should follow. However, when we actually apply IL algorithms on large datasets, we sometimes see them produce manifestly incorrect estimates of the expert's policy \citep{muller2006off, codevilla, de2019causal, chaeuffernet, kuefler2017imitating}. One possible reason for this phenomenon is that empirically, we might only have access to recordings of the expert that are corrupted by \textit{temporally correlated noise} (TCN). For example, a quadcopter pilot might have been flying under persistent wind or an expert driver might have been using a car with sticky brakes. More generally, we might expect that for a variety of sequential prediction tasks, observational data might have noise that is not independently distributed across timesteps.

The downstream effect of TCN (more formally, an unobserved confounder) is temporal correlations in the recorded actions that do not have their true cause in the recorded state. Consider again our quadcopter pilot demonstrating how to fly straight on a rather windy day. If we directly fed these swerve-filled trajectories to the learner, they might learn to reproduce the deviations, producing trajectories that deviate \textit{even further} from a straight path in a test-time windy environment. At a more abstract level, these sorts of inconsistent policy estimates can result from temporal correlations between pairs of actions (e.g. the persistent wind affecting the observed heading) being reflected in the state (e.g. the quadcopter position) leading to spurious correlations between state and action that the learner might unfortunately latch onto (e.g. turning further left when on the left).

What then should we hope to learn in these confounded settings? Given we do not have access to the unobserved confounder, a reasonable choice is to ensure that we match the behavior of an expert with access to the same information we have. Put differently, we should strive to produce actions matching those proposed by an expert \textit{queried} about our current observation. While applying an interactive imitation learning algorithm like DAgger \citep{ross2011reduction} would allow us to collect a dataset uncorrupted by confounding (as we directly observe deconfounded expert actions), a queryable expert is not a realistic assumption for many domains. We therefore focus on approaches that operate on the basis of a fixed set of demonstrations. We base our algorithms on \textit{instrumental variable regression} (IVR) \citep{angrist1996identification}, a technique from econometrics for dealing with confounding in recorded data. The high-level idea of IVR is to leverage an \textit{instrument}, a source of random variation independent of the confounder, to deconfound inputs to a learning procedure via conditioning on the instrument. In dynamical systems, history can act as this source of variation, as it is unaffected by future confounding \citep{hefny2015supervised}. Our key insight is that \textbf{\textit{we can leverage past states as instruments to break the spurious correlation between states and actions caused by an unobserved confounder}}. 
\newline
\newline
The contributions of our work are three-fold:

\noindent \textbf{1. We formalize confounding in imitation learning.} We construct a structural causal model that captures confounding from temporally correlated noise. We derive a test to detect whether TCN is present in a dataset.

\noindent \textbf{2. We present a unified derivation of modern instrumental variable regression techniques.} We show how two recent extensions of the classical IVR technique share a common structure. We extend the theoretical analysis of these previous works by deriving accuracy bounds.

\noindent \textbf{3. We provide two novel algorithms to deal with confounding in imitation learning.} We build upon modern IVR to derive two algorithms that are consistent under TCN:
\begin{itemize}[noitemsep,topsep=0pt]
    \item \texttt{DoubIL} is a generative modeling approach that can use a simulator for reduced sample complexity.
    \item \texttt{ResiduIL} is a game-theoretic and simulator-free approach.
\end{itemize}
We derive performance bounds for policies produced by these algorithms under TCN. We then validate their performance on simulated control tasks. We also empirically investigate how the persistence of the confounder impacts policy performance.

\section{Related Work}
\noindent \textbf{Imitation Learning.} Broadly speaking, imitation learning approaches can be grouped into three classes: offline, online, and interactive. Our work is most similar to offline imitation learning algorithms (e.g. Behavioral Cloning \citep{pomerleau1989alvinn}) that operate purely on collected data. Unlike previous work however, we consider the effect of unobserved confounding. Our work shares the goal of interactive imitation learning algorithms (e.g. DAgger \citep{ross2011reduction}, AggreVaTe \citep{ross2014reinforcement}), in that we seek to match what the expert would do at a particular state, rather than what is in the corrupted demonstration. Importantly, we focus on matching expert actions on \textit{expert} rollouts, rather than on \textit{learner} rollouts, as one usually does in interactive IL. Because of the unobserved confounders, the recorded expert actions and the output of an expert query would not match. Because we are only focusing on expert rollouts, we do not need an interactive expert. 

\citet{zhang2020causal, kumor2021sequential} consider imitation learning through the lens of causal inference and derive a structural condition on the inputs to the learner's policy for recovering the expert's policy. Because we only consider additive TCN, we are still able to identify causal effects without satisfying this condition -- see Sec. 5.1 of \cite{pearl1995causal} for more discussion of this point. While \cite{zhang2020causal, kumor2021sequential} give general learnability conditions, we derive efficient algorithms with performance guarantees for a specific subclass of feasible problems that are of practical interest.

Lastly, we note that we focus only on matching the actions of a deterministic expert in this work -- we leave matching arbitrary expert moments \citep{swamy2021moments} to future work.

\begin{table*}[t]
\vskip 0.15in
\begin{center}
\begin{small}
\begin{tabular}{lcccr}
\toprule
 \sc{Setting} & \sc{State} & \sc{Expert Action} & \sc{Observed Action} & \sc{Confounder}\\
 \midrule
 Quadcopter Flying & Position & Intended Heading & Actual Heading & Persistent Wind \\
 \midrule
Product Pricing & Demand & Profit Margin & Price & Raw Materials Cost \\
 \midrule
ICU Treatment & Symptoms (e.g. heartburn) & Intent to Treat & Patient Treated & Comorbidity (e.g. fever) \\  \midrule
Shared Autonomy & User State & Intended Action & Executed Action & Assistance \\
\bottomrule
\end{tabular}
\end{small}
\end{center}
\vskip -0.1in
\caption{Several examples of TCN that can lead to inconsistent estimates of the expert's policy. The first was noted empirically by \citet{ng2003autonomous} and examples related to the next two rows have been observed by \citet{wright1928tariff, desautels2017prediction, soo2019describing}.  \label{intertiatable}}
\end{table*}

\noindent \textbf{Inertia Effects in Imitation Learning.} Several authors have empirically observed a latching behavior in policies trained via imitation learning, where learned policies tend to inappropriately repeat the same action \citep{muller2006off, codevilla, de2019causal, chaeuffernet, kuefler2017imitating}. We seek to provide a plausible explanation and correction for the phenomenon reported in these works. We note that when attempting to explain these sorts of \textit{inertia effects}, \citet{de2019causal} propose causal confounding as the root cause of the learner's error. However, as pointed out by \citet{spencer2021feedback}, there is no actual confound in the theoretical or empirical examples of the work of \citet{de2019causal}, merely a high degree of covariate shift. This is because the learner observes all of the variables that influenced the expert action. We instead consider the setting with unobserved TCN.

\noindent \textbf{Instrumental Variable Regression.} The classical approach to instrumental variable regression \citep{wright1928tariff} is a two-stage least squares procedure (e.g. in \citet{angrist1996identification}'s textbook). We focus on the more general nonlinear setting and instead base our approaches on the more recent \textsc{DeepIV} \citep{hartford17a} and \textsc{AGMM} \citep{dikkala2020minimax}. We present extensions to the work in these papers, including a unified derivation of both methods and error analysis for \textsc{DeepIV}. Prior work \cite{bradtke1996linear, hefny2015supervised, chen2021instrumental} has considered using past states as an instrument for reinforcement learning. We instead focus on imitation learning and derive algorithms with policy performance bounds that factor in the strength of the past state instrument.

\section{A Brief Review of Instruments in Causal Modeling}
\label{sec:ivr}
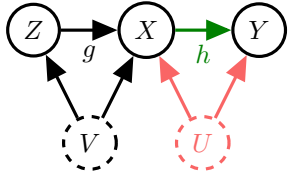
\begin{figure}[h]
    \centering
\begin{tikzpicture}[scale=1, transform shape]
    \node (a) [draw, very thick, circle,] at (0.0, 0) {$Z$};
    \node (b) [draw, very thick, circle,] at (1.5, 0) {$X$};
    \node (c) [draw, very thick, circle,]  at (3, 0) {$Y$};
    \node (d) [draw, very thick, circle, dashed, color=error]  at (2.25, -1.5) {$U$};
    \node (e) [draw, very thick, circle, dashed]  at (0.75, -1.5) {$V$};
    \path[->, color=black] (a) to[bend right] node[midway] {$g$} (b);
    \draw [->, very thick] (a) to (b);
    \path[->, color=expert] (b) to[bend right] node[midway] {$h$} (c);
    \draw [->, very thick, color=expert] (b) to (c);
    \draw [->, very thick, color=error] (d) to (b);
    \draw [->, very thick, color=error] (d) to (c);
    \draw [->, very thick,] (e) to (b);
    \draw [->, very thick,] (e) to (a);
    \end{tikzpicture}
    \caption{The structural causal model (SCM) considered in IVR. We are interested in finding $h$, the causal relationship from $X$ to $Y$, even though there is an unobserved confounder, $U$. We do so by leveraging the effect of $Z$, which provides randomness independent of $U$. \label{fig:iv}}
\end{figure} 
We begin by discussing the concept of an instrument. Let $X$, $Y$, and $Z$ be random variables on (potentially infinite) sample spaces $\mathcal{X}$, $\mathcal{Y}$, and $\mathcal{Z}$. Assume that $X$, $Y$, and $Z$ have the causal, rather than statistical, dependency structure in Fig.~\ref{fig:iv}. Given a dataset of $(x, y, z)$ tuples, we are interested in determining the causal relationship between $X$ and $Y$, $\mathbb{E}[Y|do(x)]$, where $do(\cdot)$ is the interventional operator of \citet{pearl2016causal}. Intuitively, $\mathbb{E}[Y|do(x)]$ is the expected value of $Y$ when we \textit{intervene} and set $X=x$, rather than observe such an $X$. In Fig.~\ref{fig:iv}, $h(x) = \mathbb{E}[Y|do(x)]$. Because of the presence of an unobserved confounder, $U$, that affects both $X$ and $Y$, standard regression (e.g. Ordinary Least Squares or OLS) generically produces inconsistent estimates. Coarsely, this occurs because OLS will over-estimate the influence of the parts of $X$ that are affected by the confounder. If we only have observational data or are unable to perform randomized control trials, a canonical technique to recover $h$ is IVR \citep{wright1928tariff, angrist1996identification, winship1999estimation}. Formally, an \textit{instrument} $Z$ must satisfy three structural conditions:
\vspace{-4pt}
\begin{enumerate}
    \item \textit{Unconfounded Instrument}: $Z \indep U$ -- i.e. independent randomization from confounder. \vspace{-4pt}
    \item \textit{Exclusion}: $Z \indep Y | X, U$ -- i.e. no extraneous paths. \vspace{-4pt}
    \item \textit{Relevance}: $Z \not\!\perp\!\!\!\perp X$ -- i.e. conditioning has an effect. \vspace{-4pt}
\end{enumerate}
\vspace{-4pt}
$Z$ satisfies these three conditions in the SCM of \figref{fig:iv}. \footnote{The inclusion of $V$ makes our model a generalization of the standard IVR model, so we confirm the validity of the instrument in Appendix \ref{app:proofs}.} Without loss of generality, we assume that $\mathbb{E}[U] = 0$. We further assume that noise $U$ enters additively to $Y$,\footnote{Without this assumption, one can only upper/lower bound $h(x)$ \citep{balke2013counterfactual}.} and write out the following equations:
\begin{equation}
    X = g(Z, U, V), \quad Y = h(X) + U.
    \vspace{-4pt}
\end{equation} We can now concisely derive a set of \textit{conditional moment restrictions} (CMR): 
\begin{align}
     0 = \mathbb{E}[U] = \mathbb{E}[U|z] = \mathbb{E}[Y - h(X)|z]
\label{eq:cmr} \\
    \Rightarrow \forall z \in \mathcal{Z},\, \mathbb{E}[Y|z] = \mathbb{E}[h(X)|z].
\end{align}
In words, these constraints are saying that a necessary condition for recovery of $h(x)$ is that for all values of $Z$, the actual and predicted expected values of $Y|z$ are equal. 

How can we find a predictor that satisfies the CMR? Let us first consider the setting with linear relationships between all variables. Then, one can recover $h(x) = \beta x$ by computing $\beta = \mathbb{E}[ZY]/\mathbb{E}[ZX]$. This is equivalent to the Two-Stage Ordinary Least Squares (2SLS) procedure \cite{angrist1996identification}, in which one first regresses from $Z$ to $X$ and then regresses from the predicted $\hat{X}$ to $Y$, returning the latter coefficients. Intuitively, the first stage of this procedure is aggregating $X$s based on some $z \in \mathcal{Z}$ so that the particular instantiation of $U$ that was correlating $X$ and $Y$ in the observational data has its effect ``washed out" in the $\hat{X}$s. Thus, regression from $\hat{X}$ to $Y$ is consistent.

For the more general, nonlinear problem, we can derive an appropriate loss function for finding an $\widehat{h}$ that approximately satisfies the CMR. If we have finite samples and can therefore only estimate conditional expectations up to some tolerance, it is natural to relax the CMR to
\begin{equation}
\label{eq:tik}
\begin{array}{ll@{}ll}
\min_{\widehat{h} \in \mathcal{H},\, \boldsymbol{\delta}}  &  \frac{1}{2}\mathbb{E}_z[\delta_z^2]         \\
\text{s.t.} & |\mathop{{}\mathbb{E}}[Y - \widehat{h}(X)|z]| \leq \delta_z, \quad \delta_z \geq 0, \quad \forall z \in \mathcal{Z},\\
\end{array}
\end{equation}
where the $\delta_z$ are slack variables. Then, the Lagrangian (with the natural $P(z)$-weighted inner product that captures how often each we expect each $z$ to occur) is
\begin{equation}
    L(\widehat{h}, \boldsymbol{\delta}, \boldsymbol{\lambda}) = \sum_{z \in \mathcal{Z}} P(z)\lambda_z(\mathbb{E}[Y - \widehat{h}(X)|z] - \delta_z) + P(z)\frac{1}{2}\delta_z^2,
\end{equation}
where $\boldsymbol{\lambda}$ is the vector of Lagrange multipliers. By the stationarity component of the KKT conditions, we know that
\begin{equation}
    \nabla_{\delta_z} L(\widehat{h}, \boldsymbol{\delta}, \boldsymbol{\lambda}) = -P(z)\lambda_z + P(z)\delta_z = 0,
\end{equation}
implying that $\delta_z = \lambda_z$. Plugging this back in, we can simplify the Lagrangian to
\begin{equation}
\label{eq:rela}
    L(\widehat{h}, \boldsymbol{\lambda}) = \sum_{z \in \mathcal{Z}} P(z)\lambda_z\mathbb{E}[Y - \widehat{h}(X)|z] - P(z)\frac{1}{2}\lambda_z^2.
\end{equation}
We refer to \eref{eq:rela} as the \textit{Regularized Lagrangian}. Now, solving for the optimal Lagrange multipliers via stationarity, we arrive at the expression
\begin{equation}
    \nabla_{\lambda_z} L(\widehat{h}, \boldsymbol{\lambda}) = P(z)\mathbb{E}[Y - \widehat{h}(X)|z] - P(z)\lambda_z = 0,
\end{equation}
which implies that the optimal $\lambda_z$ is equal to $\mathbb{E}[Y - \widehat{h}(X)|z]$. Plugging this back into \eref{eq:rela} produces the loss function
\begin{equation}
\label{eq:chen}
    L(\widehat{h}) = \sum_{z \in \mathcal{Z}} P(z) \mathbb{E}[Y - \widehat{h}(X)|z]^2 = \text{PRMSE}^2(\widehat{h}).
\end{equation}
This expression is the square of the \textit{Projected Root Mean Squared Error} (PRMSE) of \citet{chen2012estimation}. To recap, by minimizing \eref{eq:chen}, we are attempting to find an $\widehat{h}$ that approximately satisfies the CMR. Minimizing PRMSE is a necessary condition for recovering $\mathbb{E}[Y|do(X)]$. For it to be a sufficient condition, one needs the natural identifiability assumptions -- we refer interested readers to \citet{chen2012estimation} for a more thorough discussion.


\subsection{Generative Modeling Approach}
How should we minimize the PRMSE then? One option is learning the distribution $P(X|z) = g(z)$, passing samples from it to a candidate $\widehat{h}$, and trying to match $\mathbb{E}[Y|z]$. This is a generalization of the 2SLS procedure to nonlinear functions. The nonlinearity of the second stage means that one cannot simply compute the first moment of the $P(X|z)$ distribution, which is recovered by linearly regressing from $X$ to $Z$ in the 2SLS procedure. One instead needs to learn the entire $g(z) = P(X|z)$.
Such an approach was first proposed by \citet{hartford17a}, and amounts to first learning a $g(z)$ (e.g. via maximum likelihood estimation) and then solving
\begin{equation}
\label{eq:double_sample}
\min_{\widehat{h} \in \mathcal{H}}\mathbb{E}_Z \left[(\mathbb{E}[Y|z] - \mathbb{E}_{\hat{X} \sim g(z)}[\widehat{h}(\hat{X})])^2 \right].
\end{equation}
We note that this approach suffers from a ``double-sample" issue \citep{baird1995residual} where multiple independent samples of $g(z)$ are required to compute gradients of $\widehat{h}$. To see this, note that the gradient with respect to $\widehat{h}$ of \eref{eq:double_sample} is
\begin{equation}
    \mathbb{E}_Z\left[(\mathbb{E}[Y|z] - \mathbb{E}_{\hat{x} \sim \hat{g}(z)}[\widehat{h}(\hat{x})])(- \mathbb{E}_{\hat{x} \sim \hat{g}(z)}[\frac{\partial}{\partial \widehat{h}}\widehat{h}(\hat{x})]) \right].
\end{equation}
 Notice that $\hat{x}$ appears under two \textit{separate} expectations that are then multiplied together. Therefore, to get an unbiased estimate of this product, a minimum of two samples of $\hat{x}$ are required, one for each expectation.  
 
 The work of \citet{hartford17a} did not have theoretical analysis regarding the effect of errors in a learned $g(z)$ upon attempts to learn $h(x)$. We prove the following error bound in Appendix \ref{app:proofs}:
\begin{theorem}
Assume we learn a $g(z)$ s.t.
\begin{equation}
    \max_{\widehat{h} \in \mathcal{H}} \mathbb{E}_Z[(\mathbb{E}_{x \sim g(z)}[\widehat{h}(x)] - \mathbb{E}_{x \sim P(X|z)}[\widehat{h}(x)])^2] \leq \delta.
\end{equation}
 Then, optimizing \eref{eq:double_sample} to value $\epsilon$ corresponds to recovering a $\widehat{h}(x)$ s.t. $\text{PRMSE}(\widehat{h}) \leq \sqrt{\delta} + \sqrt{\epsilon}$. \label{thm:deepiv}
\end{theorem}


\subsection{Game-Theoretic Approach}
One can also proceed by instead solving the two-player zero-sum game with the Regularized Lagrangian \eref{eq:rela} as the payoff function. Denoting by $f \in \mathcal{F} = \{\mathcal{Z} \rightarrow \mathbb{R}\}$ the function that maps $z$'s to their Lagrange multipliers, we can write this game as 
\begin{equation}
\label{eq:rela_game}
\min_{\widehat{h} \in \mathcal{H}} \max_{f \in \mathcal{F}}  \mathbb{E}[2(Y - \widehat{h}(X))f(Z) - f(Z)^2].
\end{equation}
This game is the core objective of the AGMM method of \citet{dikkala2020minimax}. Importantly, one does not need to learn a generative model of $P(X|z)$ for these sorts of game-theoretic approaches. We prove the following theorem in Appendix \ref{app:proofs}:
\begin{theorem}
\setlength\lineskip{0pt}
Assume that $\mathcal{H}$ and $\mathcal{F}$ are bounded, closed under negation, convex, compact, $h \in \mathcal{H}$, and that $\forall \widehat{h} \in \mathcal{H}$, $f(z) = \mathbb{E}[Y-\widehat{h}(X)|z] \in \mathcal{F}$. Then, an $\epsilon$-approximate Nash equilibrium of \eref{eq:rela_game} corresponds to recovering a $\widehat{h}(x)$ s.t. $\text{PRMSE}(\widehat{h}) \leq \sqrt{\epsilon}$. \label{thm:agmm} 
\end{theorem}

One can find such an equilibrium via a standard reduction to no-regret online learning \citep{freund1997decision}. 

In summary, one can frame nonlinear IVR as a generative modeling or game-theoretic problem, leading to different error characteristics. We now turn our attention to applying these methods to imitation learning with unobserved confounders.

\section{Causal Confounding in Imitation Learning}

\begin{figure}[h]
    \centering
\begin{tikzpicture}[scale=0.9, transform shape]
    \node (a) [draw, very thick, circle] at (0.0, 0) {$a_{t-1}$};
    \node (b) [draw, very thick, circle, minimum size=1cm] at (2, 0) {$a_{t}$};
    \node (d) [circle]  at (3, 1) {$\ldots$};
    \node (n) [circle]  at (-1, 1) {$\ldots$};
    \node (e) [draw, very thick, circle] at (0.0, 2) {$s_{t-1}$};
    \node (f) [draw, very thick, circle, minimum size=1cm] at (2, 2) {$s_{t}$};
    
    \node (h) [draw, very thick, circle, dashed, color=error] at (1.0, -1.5) {$u_{t-1}$};
    \node (i) [draw, very thick, circle, dashed, minimum size=1cm] at (3, -1.5) {$u_{t}$};
    \node (o) [draw, very thick, circle, dashed, minimum size=1cm] at (-1, -1.5) {$u_{t-2}$};
    
    \node (k) [very thick, circle] at (-0.75, 1) {};
    \node (l) [very thick, circle] at (1.25, 1) {};
    \node (m) [very thick, circle]  at (3.25, 1) {};
    
    \node (p) [circle]  at (3, 2) {$X$};
    \node (q) [circle]  at (3, 0) {$Y$};
    \node (q) [circle]  at (-1, 2) {$Z$};
    

    \draw [->, very thick,] (e) to (f);
    \draw [->, very thick, color=error] (h) to (a);
    \draw [->, very thick, color=error] (h) to (b);
    \draw [->, very thick,] (i) to (b);
    \draw [->, very thick,] (o) to (a);
    \draw [->, very thick, color=expert] (e) to (a);
    \path[->, color=expert] (e) to[bend right] node[midway] {$\pi_E$} (a);
    \draw [->, very thick, color=expert] (f) to (b);
    \path[->, color=expert] (f) to[bend right] node[midway] {$\pi_E$} (b);
    \draw [->, very thick, color=error] (a) to (f);
    \path[->, ] (a) to[bend left] node[midway] {$\mathcal{T}$} (f);
    
    %
    \end{tikzpicture}
    \caption{An SCM that captures TCN. The confounding ($U = u_{t-1}$) is mediated via the dynamics into the state, introducing spurious correlations between states ($X = s_t$) and actions ($Y = a_t$). To break the confounding, we can utilize the past state as an instrument ($Z = s_{t-1}$). \label{fig:il_confounding}}
\end{figure}
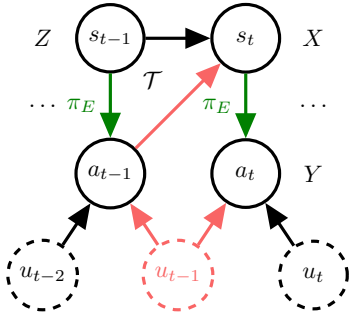

Let us introduce some IL-specific notation. We use $\Delta(S)$ to mean the set of distributions over $S$ and focus on a Markov Decision Process (MDP) parameterized by $\langle \mathcal{S}, \mathcal{A}, \mathcal{T}, r, T \rangle$, where $\mathcal{S}$ is the state space, $\mathcal{A}$ is the action space, $\mathcal{T}: \mathcal{S} \times \mathcal{A} \rightarrow \Delta(\mathcal{S})$ is the transition operator, $r: \mathcal{S} \times \mathcal{A} \rightarrow [-1, 1]$ is the reward function, and $T$ is the horizon of the problem. Let $J(\pi) = \mathop{{}\mathbb{E}}_{\tau \sim \pi}[\sum_{t=1}^T r(s_t, a_t)]$ denote the \textit{value} of policy $\pi$, $\Pi \subseteq \{\mathcal{S} \rightarrow \Delta(\mathcal{A})\}$ be the policy class we optimize over and $d_{\pi}$ be the visitation distribution of policy $\pi$. In the presence of unobserved TCN, the trajectories generated by the expert can be captured by the  structural causal model (SCM) in \figref{fig:il_confounding}.

We use $u_{t-1}$ to denote the confounder at timestep $t$. See Table \ref{intertiatable} for several examples. The confounder perturbs the past action and travels through the dynamics to influence the current state. The same confounder also perturbs the current action, leading to spurious correlations between the recorded state and action. This correlative effect is also visible in the structural equations corresponding to \figref{fig:il_confounding},
\begin{align}
    X &= s_t \\ &= \mathcal{T}(s_{t-1}, a_{t-1}) \\ &=\mathcal{T}(s_{t-1}, \pi_E(s_{t-1}) + {\color{error} u_{t-1}} + u_{t-2}) \\
    Y &= a_t = \pi_E(s_t) + u_t + {\color{error} u_{t-1}}.
\end{align}
Note the shared red term between input $X$ and output $Y$. \figref{fig:il_confounding} also tells us that $Z = s_{t-1}$ satisfies the three conditions to make it a valid instrument for countering the effects of $U = u_{t-1}$. Intuitively, the past state is independent of the current confounder, allowing it to function as an independent source of randomness. One can imagine longer time-scale correlations between actions than just the one-step connection in \figref{fig:il_confounding} -- our approaches naturally extend to this setting by using a state further back in the past as the instrument. However, this also means that the past state instrument is less predictive of the current state -- we discuss the implications of this point further in \sref{sec:exp}.

\section{What would the Expert $do(\cdot)$?: Algorithms for Causal Imitation Learning}
Essentially, standard imitation learning approaches like behavior cloning attempt to match $\mathbb{E}[a|s]$, the average observed action in the data at state $s$. An approach based on IVR instead attempts to match the \textit{interventional} effect of the expert policy, $\mathbb{E}[\pi_E(s)|s]=\mathbb{E}[a|do(s)]$. Conceptually, $\mathbb{E}[a|do(s)]$ is telling us what the expert would do on average if we intervened and \textit{placed} them in state $s$. Because of the unobserved TCN, $\mathbb{E}[a|do(s)]$ differs from $\mathbb{E}[a|s]$. 

We note that $\mathbb{E}[a|do(s)]$ is the answer we would get by averaging responses from a queryable expert in interactive approaches like DAgger \citep{ross2011reduction}. However, as we are only interested in the result of queries on states from expert demonstrations, we are able to compute $\mathbb{E}[a|do(s)]$ via IVR and do not require access to a queryable expert. 

We now present two approaches for causal imitation learning that can be seen as applications of the generative modeling and game-theoretic approaches of \sref{sec:ivr}. At their core, both algorithms are attempting to minimize a PRMSE objective,
\begin{equation}
    \min_{\pi \in \Pi} \mathbb{E}_{(s, s', a') \sim d_{\pi_E}}[(\mathbb{E}[a' - \pi(s')|s])^2], \label{eq:ivr_obj}
\end{equation}
instead of the usual offline IL objective,
\begin{equation}
    \min_{\pi \in \Pi} \mathbb{E}_{(s, a) \sim d_{\pi_E}}[(a - \pi(s))^2]. \label{eq:ols_obj}
\end{equation}
Matching symbols with \sref{sec:ivr} tells us that minimizing \eqref{eq:ivr_obj} corresponds to recovering $\mathbb{E}[a|do(s)]$. We now discuss the performance implications of approximately doing so.

\subsection{From PRMSE to Performance}
For deriving performance bounds, we assume the same distribution of TCN affects the learner at test time.\footnote{Under a different noise distribution (or no noise at all), we might do better or worse than the demonstrator. For example, on a less windy day, we are likely to do better than the quadcopter pilot at flying straight. If we make an additional \textit{overlap} assumption that we see data at all parts of the state space the learner reaches under a different noise distribution, driving \eref{eq:ivr_obj} to $0$ would imply value equivalence to the expert that has its actions affected by this new TCN distribution. Thus, we learn a policy that is \textit{value-equivariant} to the expert under a change of TCN.} Our goal in this setting is therefore to eliminate the effect of the confounder so at test time we do not needlessly reproduce its effects (e.g. the increased swerving in our quadcopter example). This is why we focus on minimizing \eref{eq:ivr_obj} instead of \eref{eq:ols_obj}. We emphasize that under TCN, minimizing the \eref{eq:ols_obj} to 0 would not recover the expert's policy.

We now define two key concepts. First, let a confounder distribution $P(U)$ be c-Total Variation stable \citep{bassily2021algorithmic} if
\begin{equation}
    \norm{a - b}_2 \leq \delta \Rightarrow d_{TV}(a + U, b + U) \leq c \delta.
\end{equation}
This property is satisfied by a wide variety of distributions (e.g. for standard normal random variables, $c=1/2$). Second, in the IL setting, the measure of ill-posedness of a problem \citep{dikkala2020minimax, chen2012estimation} is
\begin{align}
    \kappa(\Pi) &= \sup_{\pi \in \Pi} \frac{\sqrt{\mathbb{E}_{s \sim d_{\pi_E}}[(\pi_E(s) - \pi(s))^2]}}{\sqrt{\mathbb{E}_{s, s', a' \sim d_{\pi_E}}[\mathbb{E}[a' - \pi(s')|s]]^2}} \\ &= \sup_{\pi \in \Pi} \frac{\text{RMSE}(\pi)}{\text{PRMSE}(\pi)}.
\end{align}
We leverage these two definitions in the following bound on policy performance.
\begin{theorem}
\label{thm:exo_perf}
Assume $P(U)$ is $c$-TV Stable temporally correlated noise, $\pi_E$ is deterministic, and let $\kappa(\Pi)$ be the measure of the ill-posedness of the problem. Then, $\text{PRMSE}(\pi) \leq \epsilon \Rightarrow J(\pi_E) - J(\pi) \leq c \kappa(\Pi) \epsilon T^2$.
\end{theorem}

We prove this statement in Appendix \ref{app:proofs}. Intuitively, $\kappa(\Pi)$ measures the strength of the past state as an instrument. To build intuition, first consider the extreme case where $s' = s$. Then, $\kappa(\Pi) = 1$. As the past state becomes a weaker instrument, $\kappa(\Pi) > 1$. Thus, if the confounding affects multiple timesteps, we would expect $\kappa(\Pi)$ to grow as one needs to reach further back in time to find a valid instrument, leading to a looser performance bound. We investigate the effect of the length of confounding on the ill-posedness of the problem empirically in \sref{sec:exp}.

\subsection{With a Simulator: \texttt{DoubIL}}
\begin{algorithm}[H]
\begin{algorithmic}
\STATE {\bfseries Input:} Dataset $\mathcal{D}_E$ of expert trajectories, Policy class $\Pi$, Simulator $\widehat{\mathcal{T}}$
\STATE {\bfseries Output:} Trained policy $\pi_{2}$
\STATE $\pi_{1} = \arg\min_{\pi \in \Pi} \mathbb{E}_{s, a \sim \mathcal{D}_E}[-\log{\pi(a|s)}]$ 
\STATE $\mathcal{D}_{IV} = \{(\widetilde{s'} \sim \widehat{\mathcal{T}}(s, \pi_{1}(s)), a') | \forall (s, a') \in \mathcal{D}_E\}$ \hfill 
\STATE $\pi_{2} = \arg\min_{\pi \in \Pi} \mathbb{E}_{s, a \sim \mathcal{D}_{IV}}[(a - \pi(s))^2]$  \hfill 
\end{algorithmic}
\caption{\texttt{DoubIL} \label{alg:doubil}}
\end{algorithm}

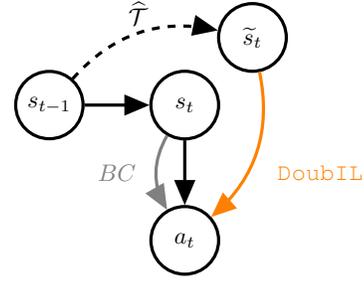
\begin{figure}[t]
    \centering
\begin{tikzpicture}[scale=0.9, transform shape]
    \node (b) [draw, very thick, circle, minimum size=1cm] at (2, 0) {$a_{t}$};
    \node (d) [circle, color=gray]  at (1, 1) {$BC$};
    \node (e) [draw, very thick, circle] at (0.0, 2) {$s_{t-1}$};
    \node (f) [draw, very thick, circle, minimum size=1cm] at (2, 2) {$s_{t}$};
    
    
    \node (k) [very thick, circle] at (-0.75, 1) {};
    \node (l) [very thick, circle] at (1.25, 1) {};
    \node (m) [very thick, circle]  at (3.25, 1) {};
    
    \node (r) [draw, very thick, circle, minimum size=1cm] at (3, 3) {$\widetilde{s}_{t}$};
    \node (s) [circle, color=orange]  at (4, 1) {$\texttt{DoubIL}$};
    
    

    \draw [->, very thick,] (e) to (f);
    \draw [->, very thick, bend left, dashed] (e) to (r);
    \path[->, color=black] (e) to[bend left] node[above] {$\widehat{\mathcal{T}}$} (r);
    \draw [->, very thick, bend left, color=orange] (r) to (b);
    \draw [->, very thick, color=black] (f) to (b);
    \draw [->, very thick, color=gray, bend right] (f) to (b);
    
    %
    \end{tikzpicture}
    \caption{\texttt{DoubIL} deconfounds inputs to the second stage regression by re-sampling state transitions via simulator $\widehat{\mathcal{T}}$. \label{fig:doubil}}
\end{figure}
Algorithm \ref{alg:doubil} can be seen as a variation of generative modeling approach of Sec. \ref{sec:ivr} and \citet{hartford17a} where one leverages knowledge of one factor of the
$P(X|z)$ distribution and just learns the other factor. Via the Markov assumption, we can factorize $P(X|z) = P(S'|s) = \sum_{a \in \mathcal{A}} P(a|s) \mathcal{T}(s, a)$. Assuming access to a simulator $\widehat{\mathcal{T}}$ that closely approximates the true transition dynamics, we can focus on learning just the $P(a|s)$ component: the standard imitation learning task. Notably, this first-stage policy is biased as it includes the effect of the confounder: $P(a|s) = P(U + \pi_E(s)|s)$. However, when we use it to simulate transitions, the next states that are drawn no longer have the particular instantiation of the confounder present in the recorded dataset’s next actions. Using a tilde to denote a fresh draw from a distribution, simulated states are drawn from
\begin{equation}
    \widetilde{s}_t \sim \widehat{\mathcal{T}}(s_{t-1}, \pi_1(s_{t-1}))
\end{equation}
while the observed next actions are drawn from
\begin{equation}
    a_t \sim \pi_E(\mathcal{T}(s_{t-1}, \pi_E(s_{t-1}) + u_{t-1} + u_{t-2})) + u_{t-1} + u_{t}.
\end{equation}
 
\begin{figure*}[t]
\centering
\begin{subfigure}{0.3\linewidth}
\centering
\includegraphics[width=\linewidth]{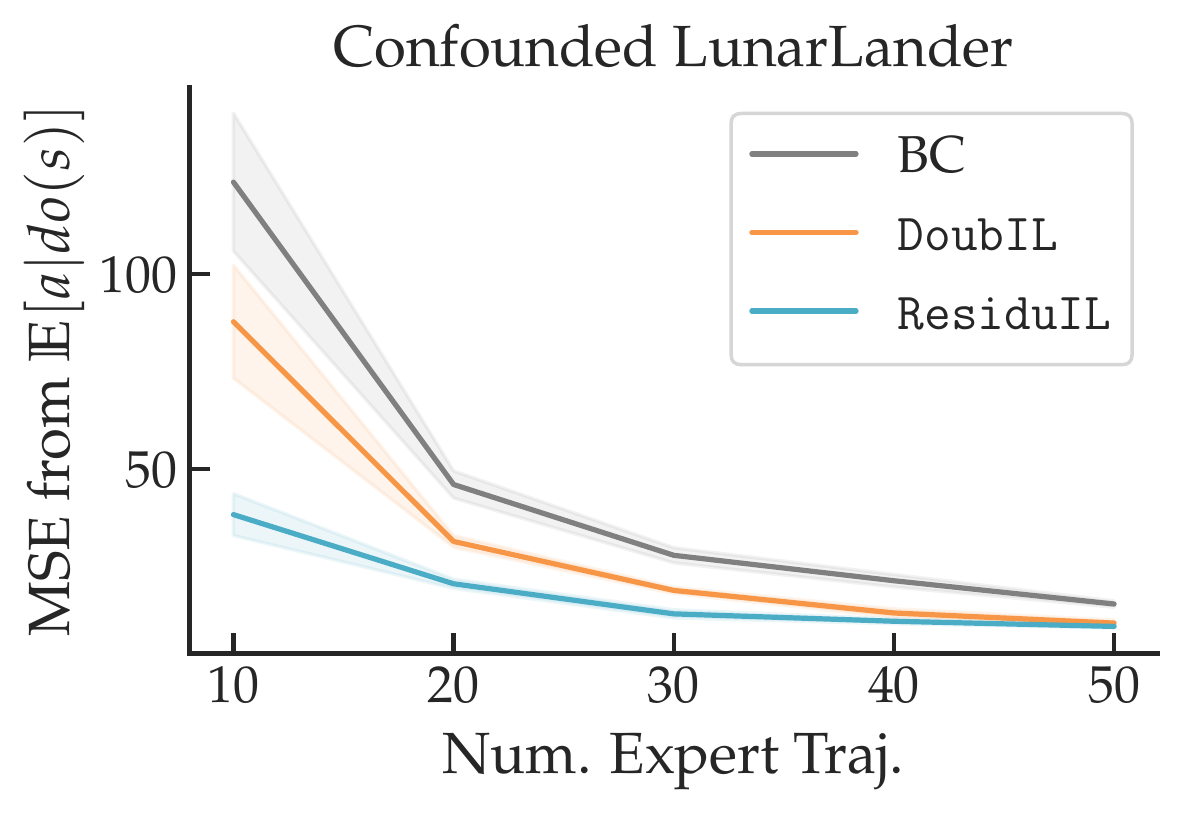}
\end{subfigure}
\begin{subfigure}{0.3\linewidth}
\centering
\includegraphics[width=\linewidth]{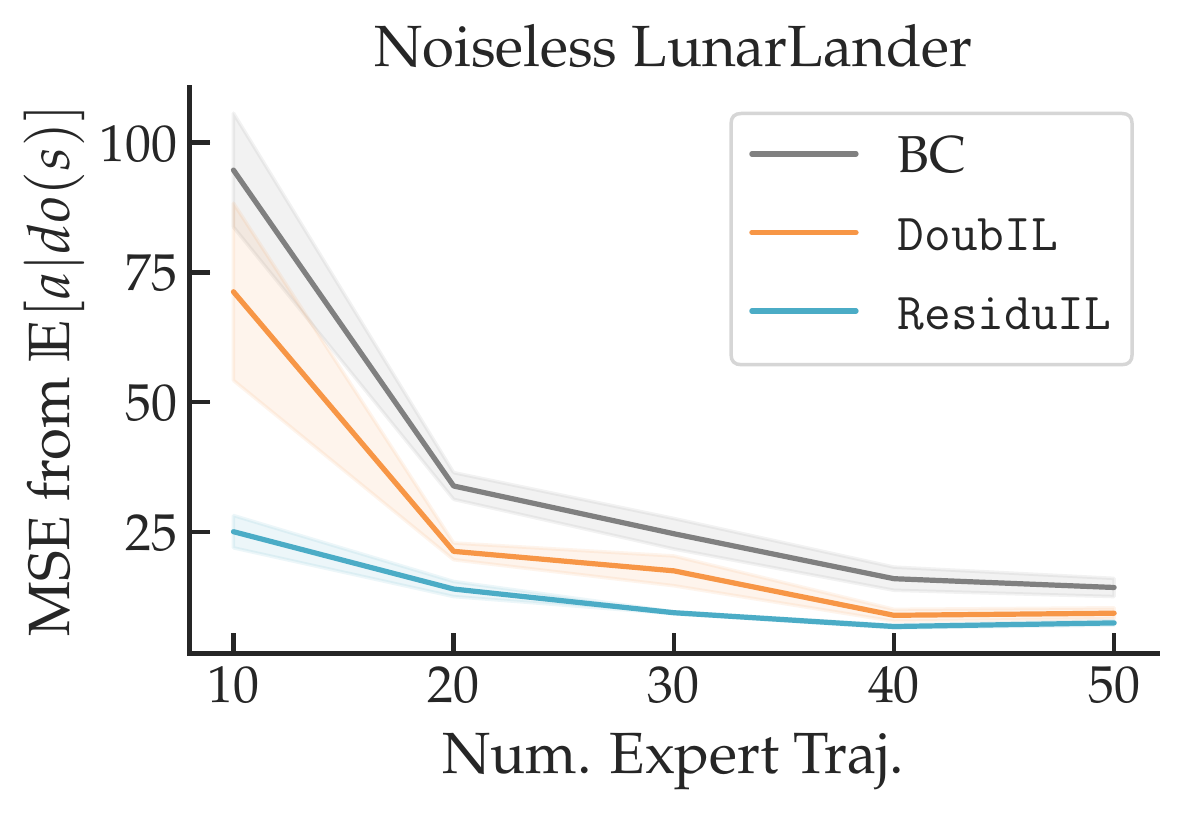}
\end{subfigure}
\begin{subfigure}{0.3\linewidth}
\centering
\includegraphics[width=\linewidth]{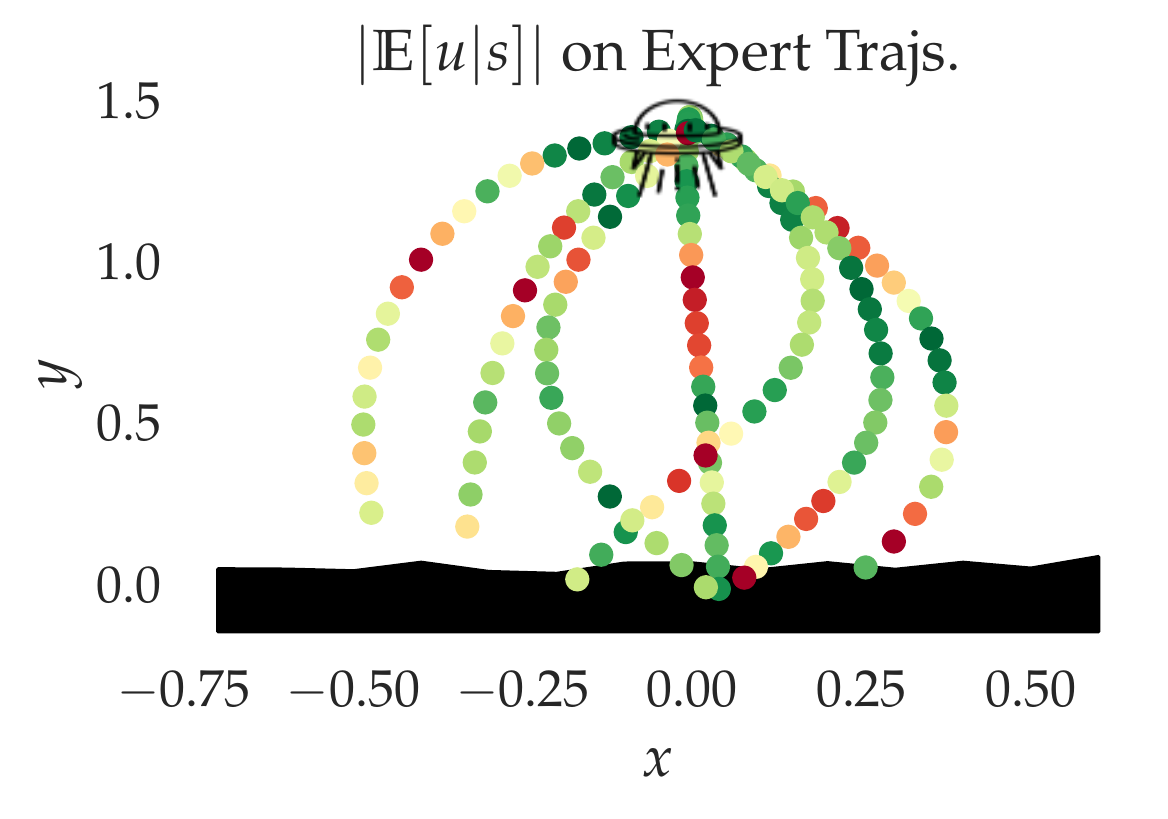}
\end{subfigure}
\caption{We train behavioral cloning, \texttt{DoubIL}, and \texttt{ResiduIL} on trajectories from a modified LunarLander environment, computing standard errors across four runs. \textbf{Left:} \texttt{DoubIL} and \texttt{ResiduIL} are better able to match $\pi_E(s) = \mathbb{E}[a|do(s)]$ on states from expert rollouts. \textbf{Center:} The policies learned by our algorithms generalize better than those produced by behavioral cloning to the state distribution of the expert on the noiseless problem ($u_t = 0$). \textbf{Right:}  We can compare the results of behavioral cloning to one of our causal IL procedures to identify areas of the state space where the effect of confounding is strong (the red dots). \label{fig:exp}}
\end{figure*}

Notice that there are no shared noise terms. This allows us to apply standard imitation learning to this new dataset of $(\widetilde{s}_t, {a}_t)$ to learn a causally consistent policy. This is because $\mathbb{E}[{a}_t|\widetilde{s}_t]=\mathbb{E}[{a}_t|do(s_t)]$. The two applications of imitation learning lead us to term this algorithm \texttt{DoubIL}. To derive a PRMSE bound, we can translate the guarantee of Theorem \ref{thm:deepiv} to our factored context:
\begin{lemma}
Assume we learn a $\pi_1(s)$ s.t. 
\begin{align}
    \max_{\pi \in \Pi} \mathbb{E}_{s_{t-1}}[(\mathbb{E}_{s_t \sim \widehat{\mathcal{T}}(s_{t-1}, \pi_1(s_{t-1}))}[\pi(s_t)] \\ - \mathbb{E}_{s_t \sim P(s_t|s_{t-1})}[\pi(s_t)])^2] \leq \delta.
\end{align}
Then, optimizing the second-stage MSE to $\epsilon$ corresponds to recovering a $\pi_{2}$ s.t. \begin{align}
    \text{PRMSE}(\pi_2) &= \sqrt{\mathbb{E}_{s \sim d_{\pi_E}}[\mathbb{E}[\pi_2(s') - \pi_E(s')|s]^2]} \\ &\leq \sqrt{\delta} + \sqrt{\epsilon}
\end{align} \label{thm:doubil}
\end{lemma}
We prove this lemma in Appendix \ref{app:proofs}. Combining this lemma with Theorem \ref{thm:exo_perf} allows one to derive a policy performance bound of 
\begin{equation}
    J(\pi_E) - J(\pi_{\texttt{DoubIL}}) \leq c \kappa(\Pi)(\sqrt{\delta} + \sqrt{\epsilon}) T^2
\end{equation}
under TCN. We note that one could simply learn the mapping $P(s'|s)$ but this can be far less sample efficient than merely learning a policy when $|\mathcal{A}| \leq |\mathcal{S}|$, as is often true in practice.
\begin{figure*}[t]
\centering
\begin{subfigure}{0.245\linewidth}
\centering
\includegraphics[width=\linewidth]{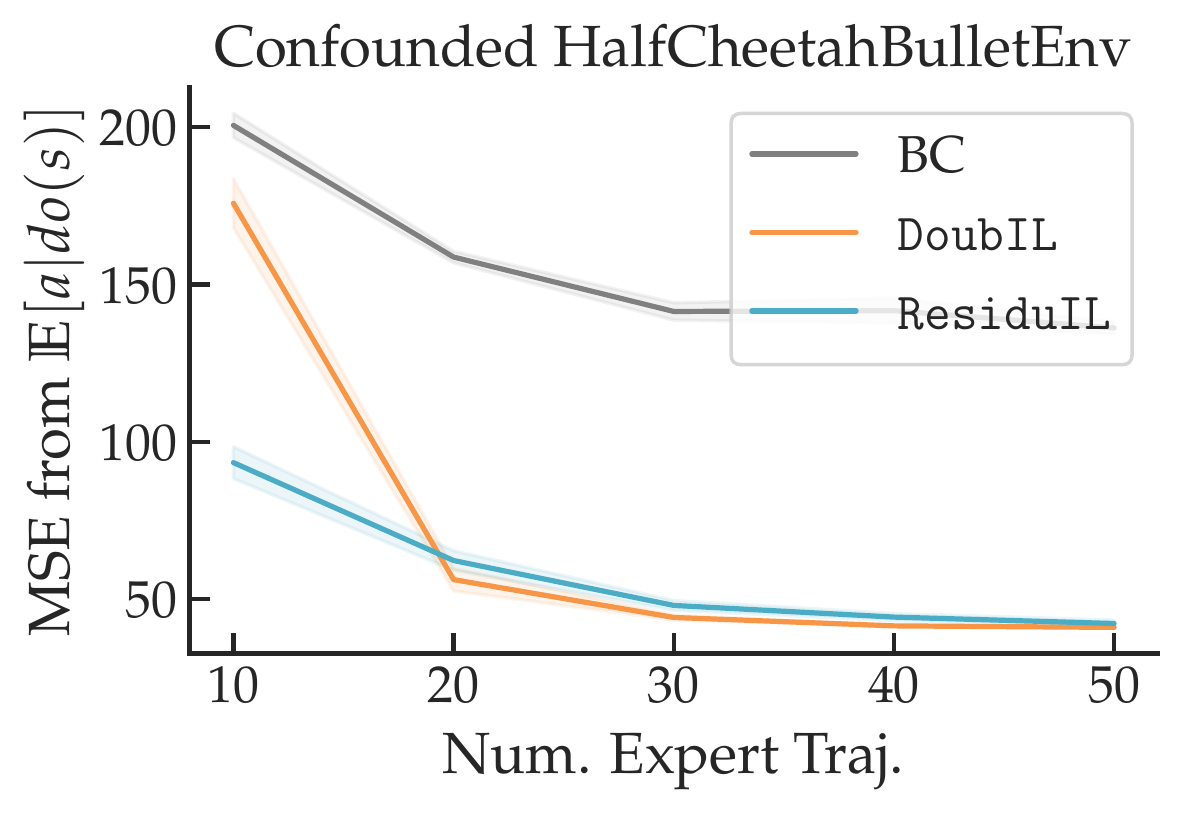}
\end{subfigure}
\begin{subfigure}{0.245\linewidth}
\centering
\includegraphics[width=\linewidth]{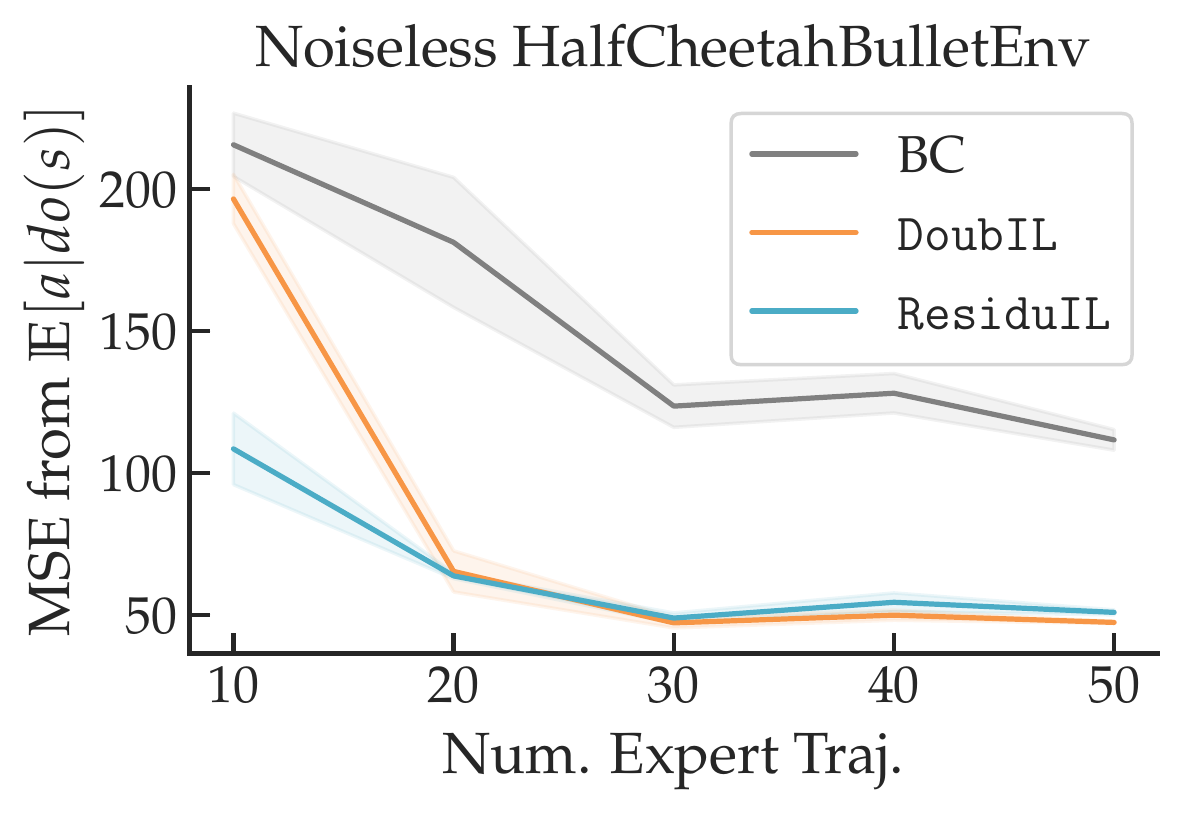}
\end{subfigure}
\begin{subfigure}{0.245\linewidth}
\centering
\includegraphics[width=\linewidth]{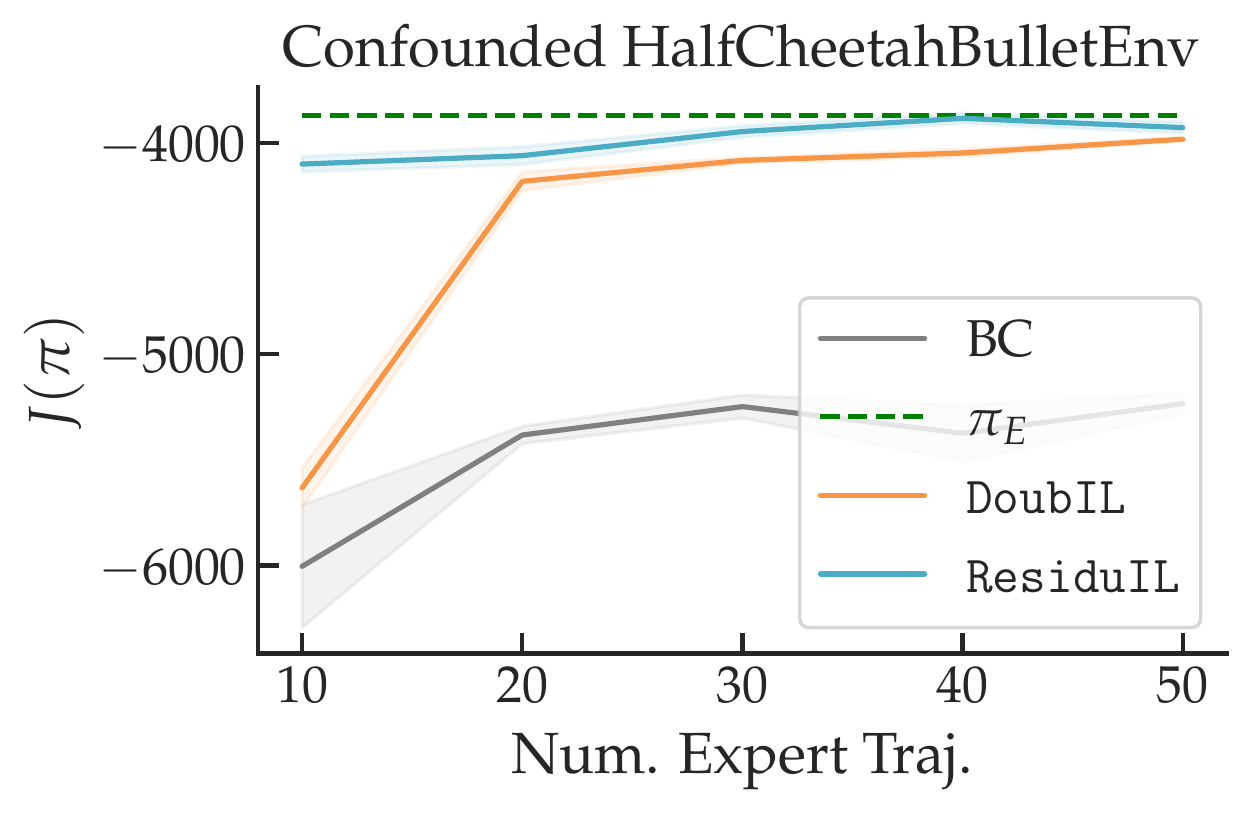}
\end{subfigure}
\begin{subfigure}{0.245\linewidth}
\centering
\includegraphics[width=\linewidth]{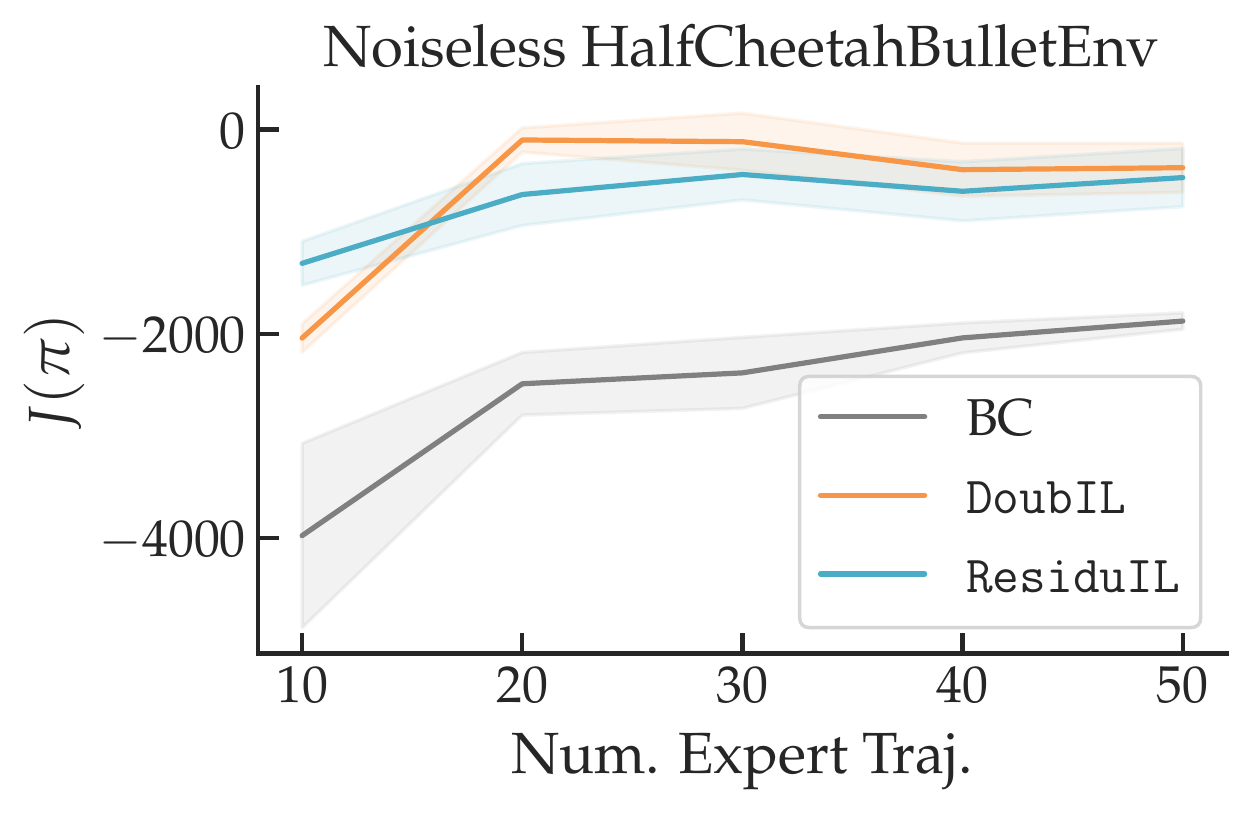}
\end{subfigure}
\begin{subfigure}{0.245\linewidth}
\centering
\includegraphics[width=\linewidth]{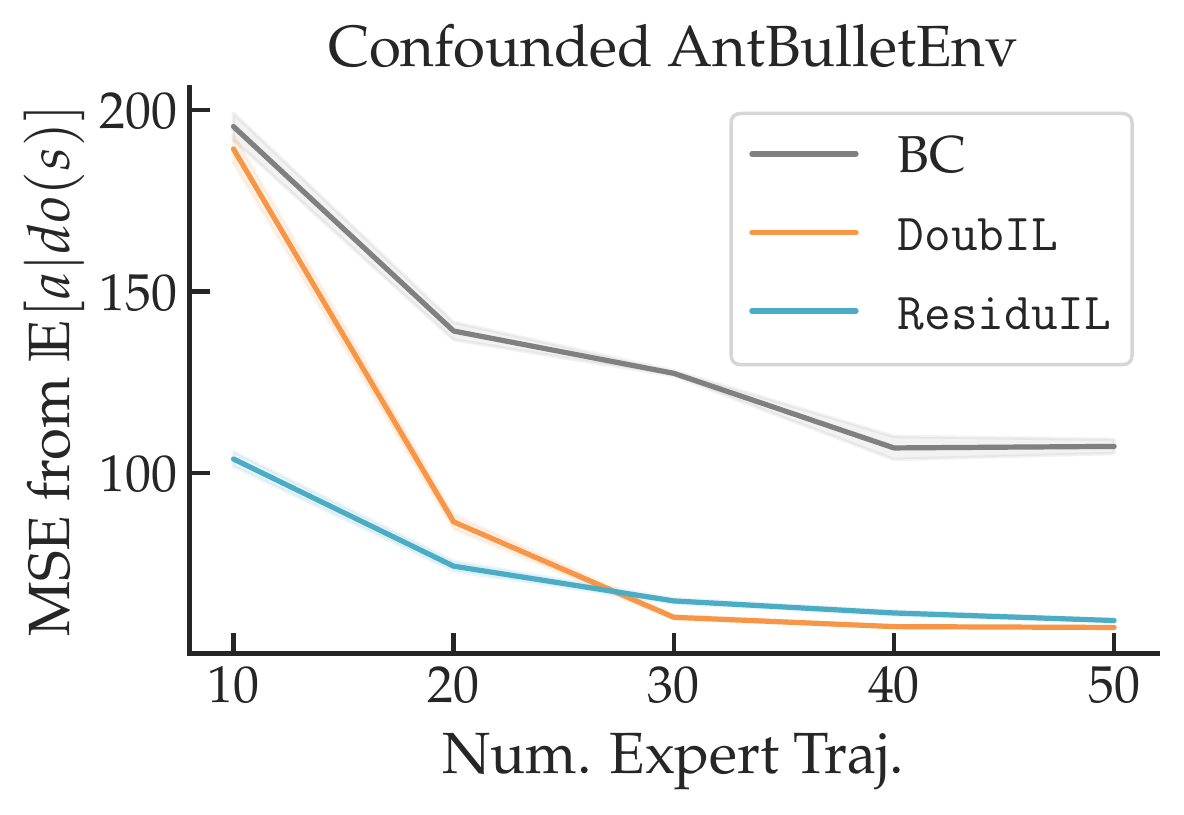}
\end{subfigure}
\begin{subfigure}{0.245\linewidth}
\centering
\includegraphics[width=\linewidth]{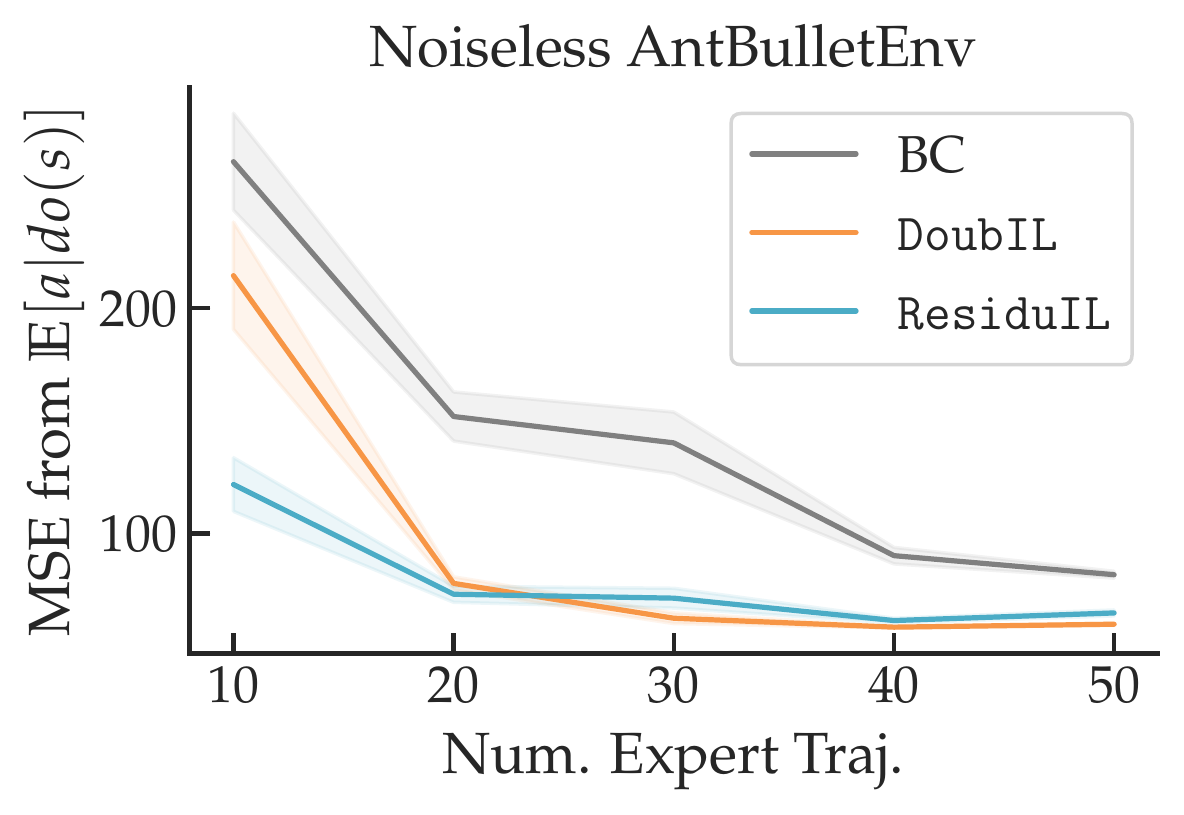}
\end{subfigure}
\begin{subfigure}{0.245\linewidth}
\centering
\includegraphics[width=\linewidth]{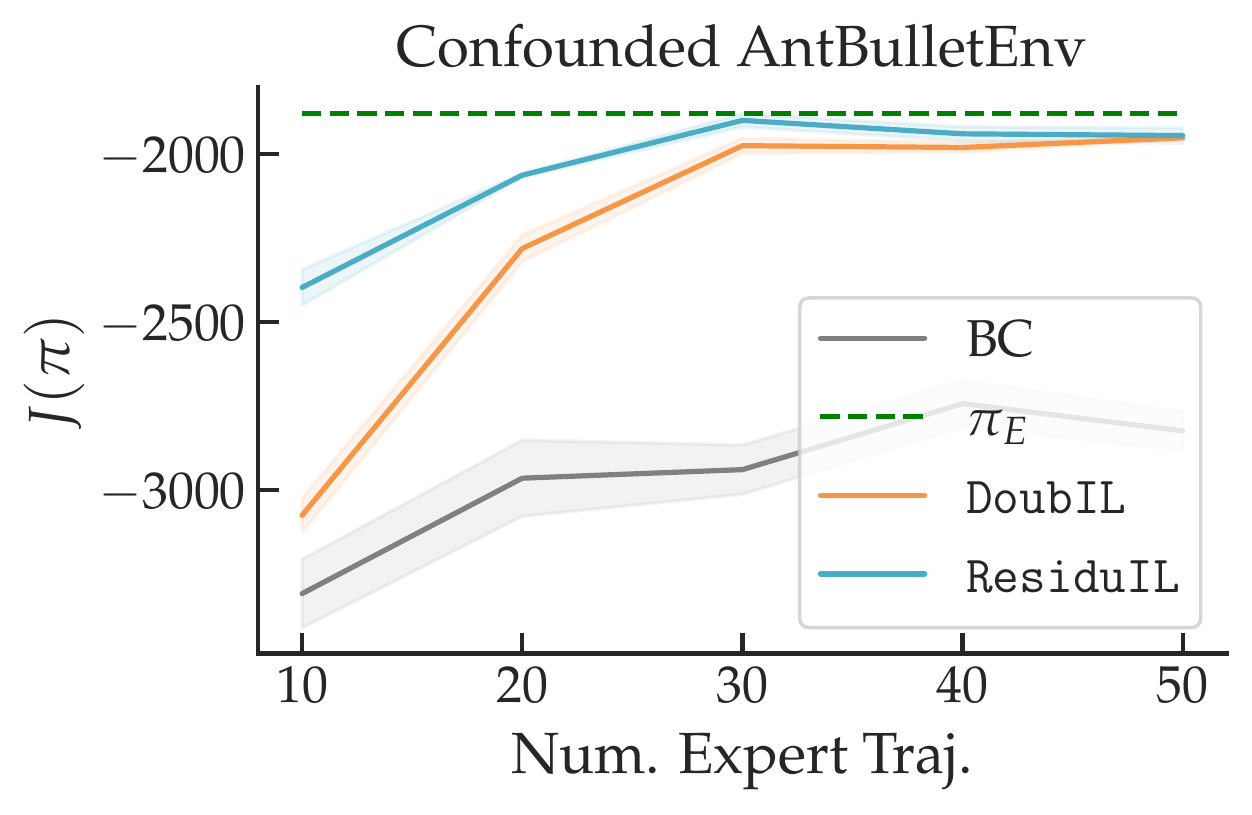}
\end{subfigure}
\begin{subfigure}{0.245\linewidth}
\centering
\includegraphics[width=\linewidth]{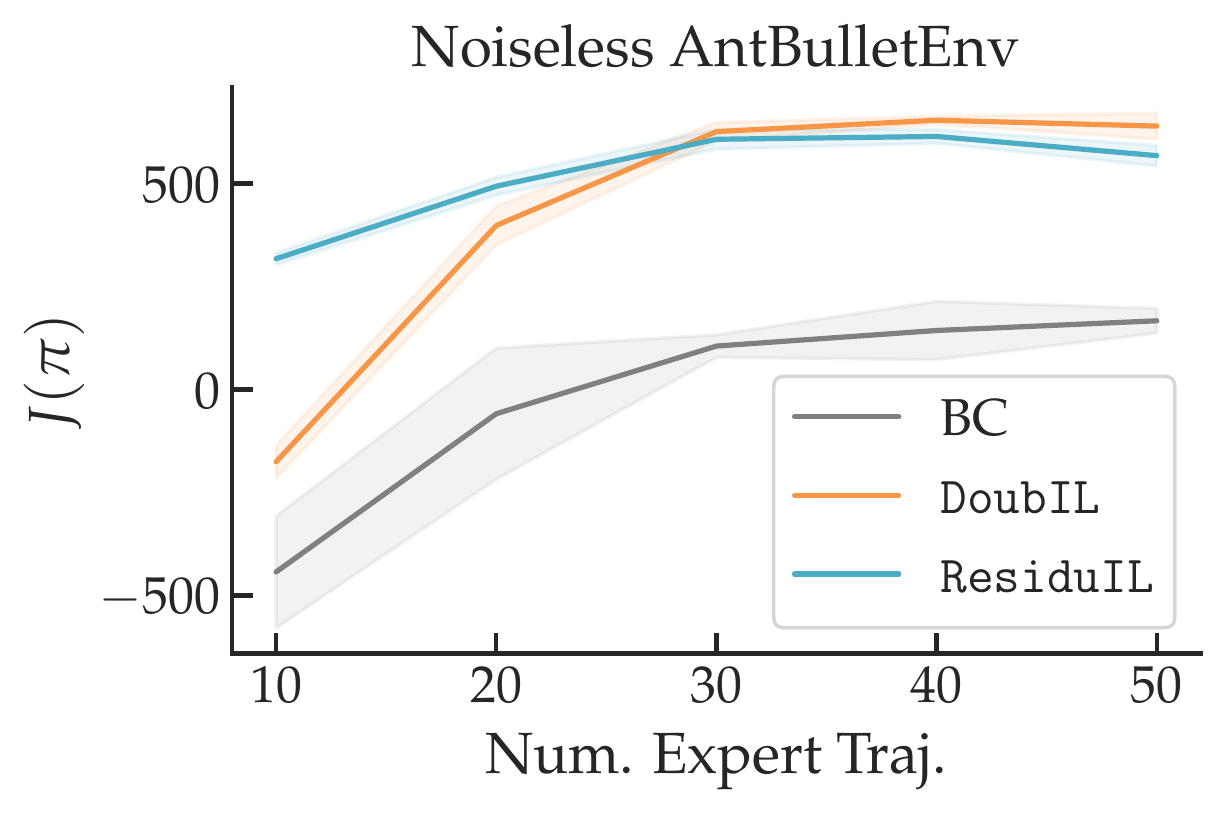}
\end{subfigure}
  \caption{We train behavioral cloning, \texttt{ResiduIL}, and \texttt{DoubIL} on trajectories from the HalfCheetahBulletEnv and AntBulletEnv environments, computing standard errors across four runs. We see both of our approaches out-perform behavioral cloning on all metrics. \label{fig:hc}}
 \end{figure*}
 
\subsection{Without state re-sampling: \texttt{ResiduIL}}
\begin{algorithm}[H]
\begin{algorithmic}
\STATE {\bfseries Input:} Dataset $\mathcal{D}_E$ of expert trajectories, Policy class $\Pi$, Discriminator class $\mathcal{F}$, Learning rate $\eta$
\STATE {\bfseries Output:} Trained policy $\pi$
\STATE Set $\pi \in \Pi$, $f \in \mathcal{F}$, $\widetilde{g}_{\pi} = 0$, $\widetilde{g}_f = 0$
\WHILE{$\pi$ not satisfactory}
\STATE $L(\pi, f) = \mathbb{E}_{(s, s', a') \sim \mathcal{D}_E}[2(a' - \pi(s'))f(s) - f(s)^2]$ 
\STATE $g_{\pi} = \nabla_{\pi} L(\pi, f)$, $g_{f} = \nabla_{f} L(\pi, f)$ 
\STATE $\pi \gets \pi - \eta(2 g_{\pi} - \widetilde{g}_{\pi})$
\STATE $f \gets f + \eta(2 g_{f} - \widetilde{g}_{f})$
\STATE $\widetilde{g}_{\pi} \gets g_{\pi}$, $\widetilde{g}_{f} \gets g_{f}$
\ENDWHILE
\end{algorithmic}
\caption{\texttt{ResiduIL} \label{alg:residuil}}
\end{algorithm}
Algorithm \ref{alg:residuil} is the direct application of the game-theoretic approach of \sref{sec:ivr} and \citet{dikkala2020minimax} to imitation learning. We term it \texttt{ResiduIL} because the adversary attempts to predict the residual between the learner and the expert's actions while the learner attempts to minimize this residual. This algorithm can be run completely offline (i.e. without access to a simulator). We use the Optimistic Mirror Descent approach of \citet{NIPS2015_7fea637f} to find approximate Nash equilibria in our experiments. Once again, we can extend our past results to the IL setting.
\begin{lemma}
An $\epsilon$-approximate equilibrium for the policy player corresponds to recovering a policy $\pi$  s.t $\text{PRMSE}(\pi) \leq \sqrt{\epsilon}$. \label{thm:residuil}
\end{lemma}
This lemma dovetails with Theorem \ref{thm:exo_perf} to prove that 
\begin{equation}
    J(\pi_E) - J(\pi_{\texttt{ResiduIL}}) \leq c \kappa(\Pi)\sqrt{\epsilon} T^2
\end{equation}
under TCN (Appendix \ref{app:proofs}). We now turn our attention to validating these guarantees empirically.

\section{Experiments}

\label{sec:exp}

 We test \texttt{DoubIL} and \texttt{ResiduIL} on a slightly modified version of the OpenAI Gym \citep{brockman2016openai} LunarLander-v2 environment against a behavioral cloning baseline. We generate demonstrations by simulating rollouts of an expert policy trained via PPO \citep{schulman2017proximal}, adding fresh Gaussian noise to the expert's action as well as cached noise from the last timestep. The latter noise is the confounder. We refer interested readers to Appendix \ref{app:exp} for full details, including hyperparameters. We see that both of our methods are able to more closely match $\mathbb{E}[a|do(s)]$ than behavioral cloning, especially in the low-data regime (\figref{fig:exp}, left). We also measure the MSE on states from deconfounded expert rollouts -- while there are no simple guarantees on this state distribution, we see that our methods generalize better than BC empirically (\figref{fig:exp}, middle). 
 
 At this point, one might wonder how, given a dataset of expert demonstrations, one detects whether there is unobserved confounding in the data. We can answer this question by comparing the results of behavioral cloning and either of our above algorithms. We prove the follow result in Appendix \ref{app:proofs}:
\begin{lemma}
Assume $\pi_{BC}(s) = \mathbb{E}[a|s]$ and $\pi_{IV}(s) = \mathbb{E}[a|do(s)]$. Then, $\mathbb{E}[u|s] = \pi_{BC}(s) - \pi_{IV}(s)$.\label{thm:test}
\end{lemma}
The implication of this lemma is that comparing the outputs of IVR-based procedures to behavioral cloning can help us detect causal confounding -- if they greatly differ with a sufficiently sized dataset, there is likely temporally correlated noise in our data. Moreover, the states where they differ represent the parts of the state space where the influence of the confounder is highest. \figref{fig:exp}, right, is an empirical example of how the test of Lemma \ref{thm:test} can be used to identify areas of the state space where the effect of the confounder is especially strong (e.g. the center).

We next consider the HalfCheetahBulletEnv and AntBulletEnv environments \cite{coumans2019}. Similar to the previous set of experiments, we train an expert via SAC \cite{haarnoja2018soft} and use Gaussian noise as the confounder -- see \cref{app:exp} for more details. In \figref{fig:hc}, We see \texttt{ResiduIL} and \texttt{DoubIL} significantly out-perform behavioral cloning across all metrics and nearly match expert performance with enough data on both confounded MDPs. This further corroborates our theory, which argues that behavioral cloning will not be able to consistently estimate the expert's policy under TCN. In contrast, our methods are able to achieve value equivalence to the expert policy.

\subsection{The Effect of TCN Persistence on Ill-Posedness}
\begin{figure}[h]
\centering
\includegraphics[width=0.6\columnwidth]{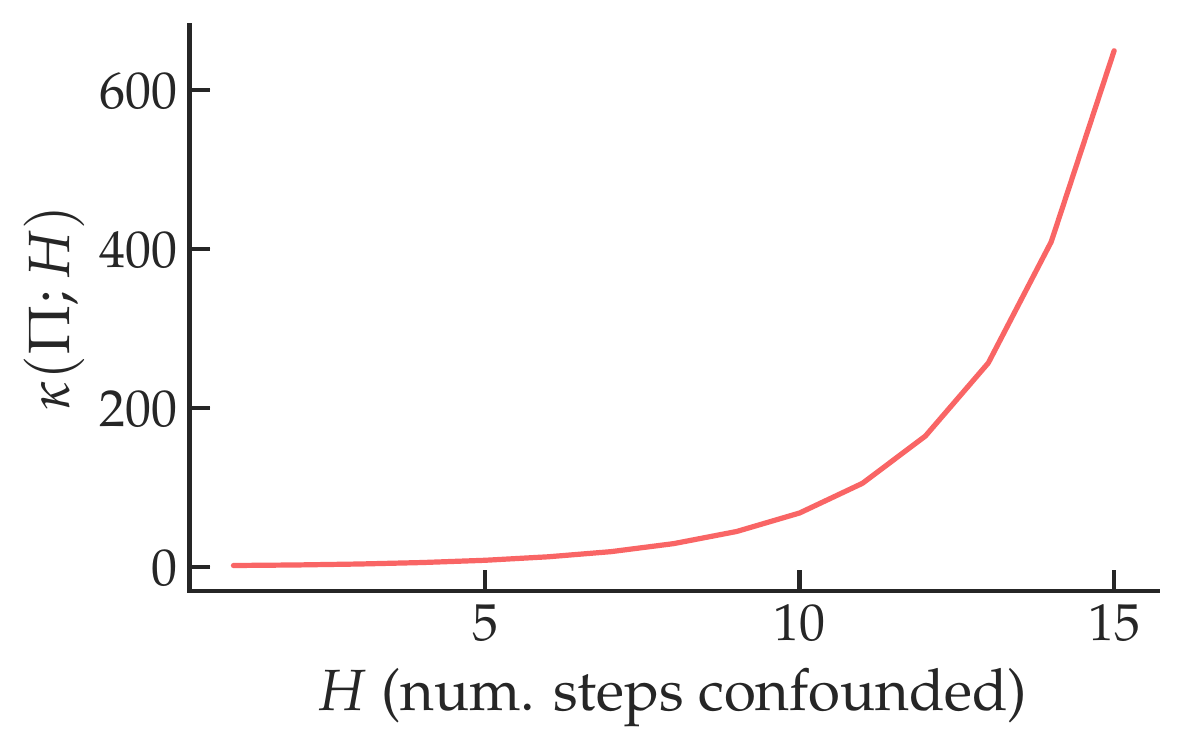}
    \caption{We compute $\kappa(\Pi)$ for an LQG problem where we vary the number of steps a confounder sticks around for. \label{fig:length}}
\end{figure}

For linear problems, we can bound $\kappa(\Pi)$ (the measure of ill-posedness) via an eigenvalue ratio \citep{dikkala2020minimax}. Extending our previous model to include the effect of the last $H$ confounders ($a_t = \pi_E(s_t) + \sum_{j = t-H}^t u_j.$), we arrive at the bound
\begin{equation}
    \kappa(\Pi; H) \leq \sqrt{\frac{\lambda_{max}(\mathbb{E}[s_{t}s_{t}^T])}{\lambda_{min}(\mathbb{E}[\mathbb{E}[s_{t}|s_{t-H}]\mathbb{E}[s_t|s_{t-H}]^T])}}. \label{eq:kappa_ub}
\end{equation}
We compute this quantity empirically for a linear-quadratic problem with Gaussian confounding and plot results in \figref{fig:length}. As expected, we see that increasing the length of confounding leads to weaker instruments as one has to use states further back in time. Theorem \ref{thm:exo_perf} therefore tells us that we should expect a larger performance gap between the learner and expert. We refer interested readers to Appendix \ref{app:exp} for full experimental setup details.

\section{Conclusion}
We present a model that captures confounding in imitation learning and derive two algorithms, \texttt{DoubIL} and \texttt{ResiduIL}, that are able to utilize history as an instrument to mitigate the effects of temporally correlated noise. We prove performance bounds and validate their empirical efficacy under TCN. We further consider how the persistence of TCN affects the performance of IVR-based imitation learning methods. We release our code at \textbf{\texttt{\url{https://github.com/gkswamy98/causal_il}}}.



\section{Acknowledgements}
GS thanks Daniel Kumor, Allie Del Giorno, Swaminathan Gurumuthy, Jonathan Spencer, and Keegan Harris for feedback on this work. ZSW is supported in part by the NSF FAI Award \#1939606, a Google Faculty Research Award, a J.P. Morgan Faculty Award, a Facebook Research Award, an Okawa Foundation Research Grant, and a Mozilla Research Grant. GS is supported by his family and friends.

\clearpage
\bibliography{example_paper}
\bibliographystyle{icml2022}

\newpage
\appendix
\onecolumn
\section{Proofs}
\label{app:proofs}
\subsection{Proof of Validity of Instrument}
\begin{proof}
We check the instrument conditions in order:
\begin{enumerate}
    \item \textit{Unconfounded Instrument}: $Z \indep U$: The $Z \rightarrow X \leftarrow U$, $V \rightarrow X \leftarrow U$, and $X \rightarrow Y \leftarrow U$ triples are blocked by standard d-separation rules \citep{pearl2016causal}. All paths from $Z$ to $U$ must pass through one of these triples so $Z \indep U$.
    \item \textit{Exclusion}: $Z \indep Y | X, U$: The $Z \rightarrow X \rightarrow Y$, $X \leftarrow U \rightarrow Y$, and $V \rightarrow X \rightarrow Y$ triples are blocked by standard d-separation rules. All paths from $Z$ to $Y$ must pass through one of these triples so $Z \indep Y | X, U$.
    \item \textit{Relevance}: $Z \not\!\perp\!\!\!\perp X$: There is a $Z \rightarrow X$ edge, which is assumed to be non-degenerate.
\end{enumerate}
Thus, $Z$ is a valid instrument for determining the causal relationship between $X$ and $Y$.
\end{proof}
\subsection{Proof of Theorem \ref{thm:deepiv}}
\begin{proof}
We simplify notation for clarity in our proof. Consider two vectors of the same dimension, $\mathbf{a}$ and $\mathbf{b}$. Assume that $\sum_i^N a_i^2 \leq \epsilon$ and $\sum_i^N b_i^2 \leq \delta$. This implies that $\norm{\mathbf{a}}_2 \leq \sqrt{\epsilon}$ and $\norm{\mathbf{b}}_2 \leq \sqrt{\delta}$. Then, by the triangle inequality, $\norm{\mathbf{a} - \mathbf{b}}_2 \leq \norm{\mathbf{a}}_2 + \norm{\mathbf{b}}_2 \leq \sqrt{\epsilon} + \sqrt{\delta}$. Setting $a_i = \sqrt{P(z)}(\mathbb{E}[Y|z] - \mathbb{E}_{\hat{x} \sim g(z)}[\widehat{h}(\hat{x})])$ and $b_i = \sqrt{P(z)}(\mathbb{E}_{\hat{x} \sim g(z)}[\widehat{h}(\hat{x})] - \mathbb{E}_{}[\widehat{h}(x)|z])$ proves that
\begin{align}
     \max_{\widehat{h} \in \mathcal{H}} \mathbb{E}_Z[(\mathbb{E}_{x \sim g(z)}[\widehat{h}(x)] - \mathbb{E}_{x \sim P(X|z)}[\widehat{h}(x)])^2] \leq \delta, \\
     \mathbb{E}_z[(\mathbb{E}[Y|z] - \mathbb{E}_{\hat{x} \sim g(z)}[\widehat{h}(\hat{x})])^2] \leq \epsilon \\
     \Rightarrow PRMSE(\widehat{h}) = \sqrt{\mathbb{E}_z[(\mathbb{E}[Y|z] - \mathbb{E}_{x \sim P(X|z)}[\widehat{h}(x)])^2]} \leq \sqrt{\epsilon} + \sqrt{\delta}
\end{align}
\end{proof}

\subsection{Proof of Theorem \ref{thm:agmm}}
\begin{proof}
Recall \eref{eq:rela_game}:
\begin{equation}
    \min_{h \in \mathcal{H}} \max_{f \in \mathcal{F}} \mathbb{E}[2(Y - h(X))f(Z) - f^2(Z)] \label{eq:pop_rela}
\end{equation}
An $\epsilon$-approximate equilibrium is an $(\widehat{h}, \widehat{f})$ pair such that:
\begin{align}
    & \max_{f \in \mathcal{F}} \mathbb{E}[2(Y - \widehat{h}(X))f(Z) - f^2(Z)] - \frac{\epsilon}{2} \\ &\leq \mathbb{E}[2(Y - \widehat{f}(X))\widehat{f}(Z) - \widehat{f}^2(Z)] \\ &\leq \min_{h \in \mathcal{H}} \mathbb{E}[2(Y - h(X))\widehat{f}(Z) - \widehat{f}^2(Z)] + \frac{\epsilon}{2} \label{eq:equib}
\end{align}
Taking the derivative w.r.t $f(z)$ of the payoff and setting it equal to 0, we arrive at
\begin{equation}
    2 P(z)\mathbb{E}[Y-\widehat{h}(X)|z] - 2 P(z) f(z) = 0 \Rightarrow f(z) = \mathbb{E}[Y-\widehat{h}(X)|z].
\end{equation}
Plugging this back into \eref{eq:equib} gives us the inequality
\begin{equation}
    \mathbb{E}_Z[\mathbb{E}[Y - \widehat{h}(X)|z]^2] - \frac{\epsilon}{2} \leq  \min_{h \in \mathcal{H}} \mathbb{E}[2(Y - h(X))\widehat{f}(Z) - \widehat{f}^2(Z)] + \frac{\epsilon}{2}.
\end{equation}
Assuming we are in the realizable setting (e.g. $h(x)=  \mathbb{E}[Y|do(x)] \in \mathcal{H}$), $\min_{h \in \mathcal{H}} \mathbb{E}[2(Y - h(X))\widehat{f}(Z) - \widehat{f}^2(Z)] \leq 0$. Thus, we can write that:
\begin{equation}
    \mathbb{E}_Z[\mathbb{E}[Y - \widehat{h}(X)|z]^2] - \frac{\epsilon}{2} \leq \frac{\epsilon}{2} \Rightarrow \text{PRMSE}(\widehat{h}) \leq \sqrt{\epsilon}.
\end{equation}
\end{proof}
We note that Theorem \ref{thm:agmm} follows somewhat directly from the main theorems of \cite{dikkala2020minimax} but that it was not stated in this precise form in their work.

\subsection{Proof of Lemma \ref{thm:doubil}}
\begin{proof}
Notice that 
\begin{equation}
    \max_{\pi \in \Pi} \mathbb{E}_{s_{t-1}}[(\mathbb{E}_{s_t \sim \widehat{\mathcal{T}}(s_{t-1}, \pi_1(s_{t-1}))}[\pi(s_t)] - \mathbb{E}_{s_t \sim P(s_t|s_{t-1})}[\pi(s_t)])^2] \leq \delta
\end{equation}
can be re-written as
\begin{equation}
     \max_{\pi \in \Pi} \mathbb{E}_Z[(\mathbb{E}_{x \sim g(z)}[\pi(x)] - \mathbb{E}_{x \sim P(X|z)}[\pi(x)])^2] \leq \delta.
\end{equation}
Thus, the proof of Theorem \ref{thm:doubil} holds as written.
\end{proof}
\subsection{Proof of Lemma \ref{thm:residuil}}
An $\epsilon$-approximate equilibrium for the policy player is a $\pi$ such that
\begin{equation}
    \max_{f \in \mathcal{F}} \mathbb{E}[2(a_{t} - \pi(s_t))f(s_{t-1}) - f^2(s_{t-1})] - \frac{\epsilon}{2} \leq \min_{\pi \in \Pi} \mathbb{E}[2(a_t - h(s_t))\widehat{f}(s_{t-1}) - \widehat{f}^2(s_{t-1})] + \frac{\epsilon}{2}. 
\end{equation}
With a change of notation, we can re-write this as:
\begin{equation}
    \max_{f \in \mathcal{F}} \mathbb{E}[2(Y - \pi(X))f(Z) - f^2(Z)] - \frac{\epsilon}{2} \leq \min_{\pi \in \Pi} \mathbb{E}[2(Y - h(X))\widehat{f}(Z) - \widehat{f}^2(Z)] + \frac{\epsilon}{2}. \label{eq:equib}
\end{equation}
Thus, the proof of Theorem \ref{thm:agmm} holds as written.
\subsection{Proof of Theorem \ref{thm:exo_perf}}
\begin{proof}
By definition,
\begin{equation}
    \text{PRMSE}(\pi) = \sqrt{\mathbb{E}_{s \sim d_{\pi_E}}[\mathbb{E}[a' - \pi(s')|s]]^2} = \epsilon.
\end{equation}
Recall that the measure of ill-posedness of the problem \citep{dikkala2020minimax, chen2012estimation} can be defined as
\begin{equation}
    \kappa(\Pi) = \sup_{\pi \in \Pi} \frac{\sqrt{\mathbb{E}_{s \sim d_{\pi_E}}[(\pi_E(s) - \pi(s))^2]}}{\sqrt{\mathbb{E}_{s, s', a' \sim d_{\pi_E}}[\mathbb{E}[a' - \pi(s')|s]]^2}} = \sup_{\pi \in \Pi} \frac{\text{RMSE}(\pi)}{\text{PRMSE}(\pi)}
\end{equation}

Directly,
\begin{equation}
    \text{RMSE}(\pi) \leq \epsilon \kappa(\Pi).
\end{equation}
We repeat the definition of total variation stability of a distribution $P(U)$:
\begin{equation}
    \norm{a - b}_2 \leq \delta \Rightarrow d_{TV}(a + U, b + U) \leq c \delta.
\end{equation}
We proceed by noting that TV-stability implies that $\forall s \in \mathcal{S}$,
\begin{equation}
    d_{TV}(\pi(s) + U, \pi_E(s) + U) \leq c  \norm{\pi(s) - \pi_E(s)}
\end{equation}
\begin{equation}
    \Rightarrow d_{TV}(\pi(s) + U, \pi_E(s) + U)^2 \leq c^2  \norm{\pi(s) - \pi_E(s)}^2
\end{equation}
\begin{equation}
    \Rightarrow \mathbb{E}_{s \sim d_{\pi_E}}[d_{TV}(\pi(s) + U, \pi_E(s) + U)^2] \leq c^2  \mathbb{E}_{s \sim d_{\pi_E}}[\norm{\pi(s) - \pi_E(s)}^2] = c^2 \text{MSE}(\pi). 
\end{equation}
By Jensen's inequality,
\begin{equation}
     \mathbb{E}_{s \sim d_{\pi_E}}[d_{TV}(\pi(s) + U, \pi_E(s) + U)] ^2 \leq \mathbb{E}_{s \sim d_{\pi_E}}[d_{TV}(\pi(s) + U, \pi_E(s) + U)^2] \leq c^2 \text{MSE}(\pi).
\end{equation}
Taking the square root of both sides, we arrive at
\begin{equation}
     \mathbb{E}_{s \sim d_{\pi_E}}[d_{TV}(\pi(s) + U, \pi_E(s) + U)] \leq c \text{ RMSE}(\pi) \leq c \kappa(\Pi) \epsilon.
\end{equation}
Lastly, we apply the Performance Difference Lemma of \cite{Kakade02approximatelyoptimal} as follows:
\begin{align}
    J(\pi_E) - J(\pi) &= T \mathbb{E}_{s, a \sim d_{\pi_E}}[Q^{\pi}(s, a) - \mathbb{E}_{a' \sim \pi(s)}[Q^{\pi}(s, a')]] \\
     &= T \mathbb{E}_{s, a \sim d_{\pi_E}}[Q^{\pi}(s, \pi_E(s) + u + \widetilde{u}_1) - \mathbb{E}[Q^{\pi}(s, \pi(s) + u + \widetilde{u}_2)]] \\
     & \leq T^2  \mathbb{E}_{s \sim d_{\pi_E}}[d_{TV}(\pi(s) + U, \pi_E(s) + U)] \\
     & \leq c \kappa(\Pi) \epsilon T^2.
\end{align}
We use the fact that the same $u$ would be added to both the learner and the expert's actions and that rewards are in the range $[-1, 1]$ in the third step.

\end{proof}
\subsection{Proof of Lemma \ref{thm:test}}
\begin{proof}
\begin{equation}
      \mathbb{E}[a_t|do(s_t)] = \mathbb{E}[\pi_E(s_t) + u_{t} + u_{t-1} | do(s_t)] = \pi_E(s_t) + \mathbb{E}[u_t] + \mathbb{E}[u_{t-1}] = \pi_E(s_t)
\end{equation}
\begin{equation}
     \mathbb{E}[a_t|s_t] = \mathbb{E}[\pi_E(s_t) + u_{t} + u_{t-1} |s_t] = \pi_E(s_t) + \mathbb{E}[u_t] + \mathbb{E}[u_{t-1}|s_t] = \pi_E(s) + \mathbb{E}[u_{t-1}|s_t]
\end{equation}
\begin{equation}
    \pi_{BC}(s) - \pi_E(s) = \mathbb{E}[a_t|s_t] - \mathbb{E}[a_t|do(s_t)] = \mathbb{E}[u_{t-1}|s_t] = \mathbb{E}[u|s]
\end{equation}
\end{proof}

\section{Experiment Details}
\label{app:exp}
\subsection{LunarLander Experiments}
For ease of simulation, we remove the legs from the LunarLander vehicle (the joints connecting them to the main body have a state that is not recorded in the observed state), remove the dispersion noise, and generate trajectories with a fixed ground layout. 

For all learned functions, we use two-layer ReLu MLPs with 256 hidden units. We use the Adam optimizer \citep{kingma2014adam} for behavioral cloning and \texttt{DoubIL} and use the optimistic variant for \texttt{ResiduIL}. We apply a weight decay of 1e-3 to all. We train all methods for 50k steps.
\begin{table}[h]
\begin{center}
\begin{small}
\begin{sc}
\setlength{\tabcolsep}{2pt}
\begin{tabular}{lccccr}
\toprule
 Parameter & Value \\
\midrule
 Learning Rate & 3e-4 \\
 Batch Size & 128 \\
\bottomrule
\end{tabular}
\end{sc}
\end{small}
\end{center}
\caption{Parameters for behavioral cloning.}
\end{table}

For computational ease, we only learn the mean of $P(a|s)$ for \texttt{DoubIL} and add fresh, appropriately scaled normal noise on-top of it to simulate drawing actions. For more complex noise models, one would need to use a moment matching algorithm \citep{swamy2021moments} in the first stage.

\begin{table}[h]
\begin{center}
\begin{small}
\begin{sc}
\setlength{\tabcolsep}{2pt}
\begin{tabular}{lccccr}
\toprule
 Parameter & Value \\
\midrule
 Learning Rate & 3e-4 \\
 Batch Size & 128 \\
 Num. Samples for $\mathbb{E}$ & 4 \\
\bottomrule
\end{tabular}
\end{sc}
\end{small}
\end{center}
\caption{Parameters for \texttt{DoubIL}.}
\end{table}

For implementing the ``double samples" for the gradient, we compute $\mathbb{E}_1[a'-\pi(s')|s]$ and $\mathbb{E}_2[a'-\pi(s')|s]$ using independent samples. Then, we apply a stop-gradient operator to the former expectation before taking a product between the expectations and averaging over $s$:
\begin{equation}
    L(\pi) = \mathbb{E}_s[{\color{red}\text{\Stopsign}}(\mathbb{E}_1[a'-\pi(s')|s]) \mathbb{E}_2[a'-\pi(s')|s]].
\end{equation}
This loss function has the correct gradient as it uses independent samples for computing the two expectations.

\begin{table}[h]
\begin{center}
\begin{small}
\begin{sc}
\setlength{\tabcolsep}{2pt}
\begin{tabular}{lccccr}
\toprule
 Parameter & Value \\
\midrule
 Learning Rate & 5e-5 \\
 Batch Size & 128 \\
 BC Regularizer Weight & 5e-2 \\
 $f$ norm Penalty & 1e-3 \\
 Adam $\beta$s & 0, 1e-2 \\
\bottomrule
\end{tabular}
\end{sc}
\end{small}
\end{center}
\caption{Parameters for \texttt{ResiduIL}.}
\end{table}

We use Gaussian noise with $\sigma=0.5$.

\subsection{PyBullet Experiments}
We increase the weight decay for all networks to 5e-3. We keep the same parameters for \texttt{DoubIL} (except for increasing the number of samples for $\mathbb{E}$ to 8 for AntBulletEnv) and \texttt{BC} as for the LunarLander experiments. We use the following parameters for \texttt{ResiduIL}.

\begin{table}[h]
\begin{center}
\begin{small}
\begin{sc}
\setlength{\tabcolsep}{2pt}
\begin{tabular}{lccccr}
\toprule
 Parameter & Value \\
\midrule
 Learning Rate & 5e-5 \\
 Batch Size & 128 \\
 BC Regularizer Weight & 0 \\
 $f$ norm Penalty & 1e-3 \\
 Adam $\beta$s & 0, 1e-2 \\
\bottomrule
\end{tabular}
\end{sc}
\end{small}
\end{center}
\caption{Parameters for \texttt{ResiduIL}.}
\end{table}

We use Gaussian noise with $\sigma = 2$ for AntBulletEnv and $\sigma = 3$ for HalfCheetahBulletEnv.

\subsection{LQG Experiments}
We compute the optimal policy for the following canonical linear system via solving a Discrete-Time Algebraic Ricatti Equation via the standard iterative method: 
\begin{equation}
x_{t} = A x_{t-1} + B u_{t-1}
\end{equation}
\begin{equation}
J(K) = \sum_t^T x_t^T Q x_t + (Kx_t)^T R Kx_t
\end{equation}
\[
A=
  \begin{bmatrix}
    1 & \Delta T  \\
    0 & 1 
  \end{bmatrix},
  B=
  \begin{bmatrix}
    0.5 (\Delta T)^2  \\
    \Delta T
  \end{bmatrix},
  Q=
  \begin{bmatrix}
    1 & 0  \\
    0 & 1 
  \end{bmatrix},
 R=
  \begin{bmatrix}
    0.1  \\
  \end{bmatrix}, \Delta T = 0.1
\]

This is the dynamics of a ``sliding brick on a frozen lake." We then simulate rollouts of 200 timesteps with $u_t$ being drawn i.i.d. from the standard normal distribution. We confound actions with the sum of confounders going $H$ steps back:
\begin{equation}
    a_t = K^* s_t + \sum_{j = t-H}^t u_j.
\end{equation}
We simulate 1000 such rollouts to compute \eref{eq:kappa_ub} empirically. We calculate $\mathbb{E}[X|z] = \mathbb{E}[s_t|s_{t-H}] = (A + BK^*)^H s_{t-H}$ analytically instead of via samples due to the small value of the quantity in comparison to the variance of the noise.
\end{document}